\documentclass[11pt]{article}

\usepackage[T1]{fontenc}
\usepackage[utf8]{inputenc}
\usepackage{lmodern}

\usepackage[a4paper,margin=1in]{geometry}

\usepackage{amsmath,amssymb,amsthm}

\newtheorem{lemma}{Lemma}
\newtheorem{proposition}{Proposition}
\newtheorem{corollary}{Corollary}

\usepackage{graphicx}
\usepackage{booktabs}
\usepackage{tabularx}
\usepackage{longtable}

\newenvironment{longtable*}{\begin{longtable}}{\end{longtable}}

\usepackage{float}              
\usepackage[section]{placeins}  

\usepackage[version=4]{mhchem}
\usepackage{siunitx}

\usepackage{algorithm}
\usepackage{algorithmic}

\usepackage[numbers,sort&compress]{natbib}
\usepackage[hidelinks]{hyperref}

\usepackage{authblk}

\newcommand{\AIAAarXivNotice}{%
  \begingroup
  \renewcommand{\thefootnote}{}%
  \footnotetext{%
    \footnotesize
    Preprint (author manuscript). Copyright \textcopyright\ 2026 by Project-S Space and Beyond Technologies UG.%
  }%
  \addtocounter{footnote}{-1}%
  \endgroup
}

\title{Towards Space-Based Environmentally-Adaptive Robotic Grasping}

\author[1,2]{Leonidas Askianakis}
\author[2]{Aleksandr Artemov}

\affil[1]{Technical University of Munich, Department of Aerospace and Geodesy, Munich, Germany}
\affil[2]{Project-S, Munich, Germany}

\date{} 

\begin{document}

\maketitle

\AIAAarXivNotice

\begin{abstract}
Robotic manipulation in unstructured environments requires reliable execution under diverse conditions, yet many state‑of‑the‑art systems still struggle with high‑dimensional action spaces, sparse rewards, and slow generalization beyond carefully curated training scenario. We study these limitations through the example of grasping in space environments. We learn control policies directly in a learned latent manifold that fuses (grammarizes) multiple modalities into a structured representation for policy decision‑making. Building on GPU‑accelerated physics simulation, we instantiate a set of single‑shot manipulation tasks and achieve over 95\% task success with Soft Actor‑Critic (SAC)‑based reinforcement learning in less than 1M environment steps, under continuously varying grasping conditions from step 1. This empirically shows faster convergence than representative state‑of‑the‑art visual baselines under the same open‑loop single‑shot conditions. Our analysis indicates that explicitly reasoning in latent space yields more sample‑efficient learning and markedly improved robustness to novel object and gripper geometries, environmental clutter, and sensor configurations compared to standard baselines. These mark a substantial addition to the space-based robotic grasping paradigm. At the same time, we identify remaining limitations and outline directions towards fully adaptive and generalizable grasping in the extreme conditions of space.
\end{abstract}


%

\section{Introduction}
\label{sec:intro}

Robotic grasping remains difficult to deploy reliably outside carefully curated laboratory settings, especially when both
(i) the \emph{target distribution} (object geometry, mass distribution, surface properties, and even the end-effector morphology) and
(ii) the \emph{operating conditions} (contact mechanics, sensing conditions, and dynamics regimes) vary across episodes.
The challenge is amplified in safety-critical deployments such as on-orbit servicing and debris remediation, where environmental effects
(e.g., thermal cycling, vacuum-driven tribology changes, radiation-induced material aging, and illumination extremes) can perturb both
sensing and contact over long mission timelines, while repeated on-hardware trial-and-error is costly or infeasible
\cite{medina2017refuel,jaekel2018edeorbit,mavrakis2021rocketstage,Crawford2019SpaceTribology,Hirschmugl2022RadiationRubber}.
A central practical requirement is therefore \emph{rapid learning}: reaching a high grasp success rate with as few environment interactions
as possible, while maintaining robustness to out-of-distribution (OOD) objects and context shifts.

\vspace{0.5em}
\noindent
\textbf{Motivation: sample efficiency and explicit environment conditioning.}
Classical analytic grasp synthesis provides strong geometric priors and interpretable stability criteria
\cite{nguyen1988forceclosure,ferrari1992optimal,cutkosky1989graspchoice,shimoga1996survey,sahbani2012overview},
but typically assumes accurate object models and contact parameters and is brittle under distribution shift.
Learning-based grasping and reinforcement learning (RL) can adapt to variability \cite{levine2016handeye,pinto2016selfsupgrasp},
yet policies trained directly from high-dimensional perceptual inputs are often sample-hungry and sensitive to implementation details
\cite{henderson2018drlmatters}.
A common mitigation is domain/dynamics randomization \cite{tobin2017domainrand,peng2018dynrand}, but randomization alone treats regime
changes as \emph{unobserved} stochasticity that the policy must implicitly average over.

This paper studies a more structured alternative: \emph{if regime descriptors are measurable or specifiable, integrate them explicitly as
policy conditioning variables}. Concretely, we consider an episode-level environment/context vector
$e \in [-1,1]^{d_e}$, which in principle may represent mission or scene descriptors (e.g., thermal or vacuum proxies),
and in this paper is instantiated in simulation as a normalized encoding of episode-wise contact and dynamics parameters
(friction, mass scaling, gravity, restitution, and damping; Sec.~\ref{sec:setup}).
This design is intended to enable \emph{zero-shot} adaptation across regimes by conditioning, without training separate policies.

\vspace{0.5em}
\noindent
\textbf{Evaluation stance: single-shot open-loop exteroception.}
Because closed-loop visual servoing can dominate performance (and changes the nature of the learning problem), we evaluate the main
representation and conditioning claims in a \emph{single-shot open-loop} setting:
the agent receives one exteroceptive snapshot at the start of the episode and then executes the grasping sequence without further
exteroceptive updates.\footnote{%
This paper uses ``open-loop'' in the sense of \emph{no online exteroceptive feedback} (e.g., no continuous RGB/RGB-D stream).
Proprioceptive state (e.g., joint configuration) may still be available to the low-level controller depending on the task definition in
Sec.~\ref{sec:problem}.%
}
This protocol isolates improvements due to representation and decision-making, rather than benefits that arise purely from online perception.
We compare against a representative \emph{one-shot visual} baseline under the same open-loop constraint (Sec.~\ref{sec:exp}).

\vspace{0.5em}
\noindent
\textbf{Approach: environmentally-adapted grammarization for fast policy learning.}
Our core technical idea is to avoid learning control directly in raw observation space and instead learn in a compact, structured latent
space that fuses heterogeneous inputs into a task-relevant \emph{grammar}.
Building on the grammarization framework \cite{askianakis2024grammarization,askianakis2024grammarizationbasedgraspingdeepmultiautoencoder,Askianakis2025},
we construct an extended latent representation
\begin{equation}
z_C \;=\; \big[z_q \,\Vert\, z_s \,\Vert\, e\big] \in \mathbb{R}^{4 + d_s + d_e},
\label{eq:zc_intro}
\end{equation}
where $z_q \in \mathbb{R}^{4}$ is a dedicated quaternion/orientation channel, $z_s \in \mathbb{R}^{d_s}$ captures task-relevant
shape/scene/gripper factors, and $e$ is the normalized context block.
A reinforcement learning policy (SAC \cite{haarnoja2018sac}) operates on $z_C$ (together with minimal geometric state, Sec.~\ref{sec:setup})
to output continuous control actions.

To keep rotation variables numerically well-behaved, we apply a deterministic unit-norm projection to obtain valid quaternions at the
simulator interface and mitigate antipodal discontinuities, consistent with best practices for rotation parameterizations
\cite{kuipers1999quaternions,zhou2019rotationcontinuity}.
We further propose a mutual-information (MI) decoupling mechanism between the orientation block and the non-orientation block
$(z_s,e)$ using an InfoNCE-style estimator with a hinge ceiling \cite{oord2018cpc,poole2019mibounds} to reduce cross-talk during learning.

\vspace{0.5em}
\noindent
\textbf{Experimental platform and headline empirical result.}
We implement the pipeline in the ManiSkill manipulation benchmark \cite{gu2023maniskill2} (SAPIEN-based simulation \cite{xiang2020sapien}),
enabling high-throughput training under episode-wise dynamics randomization applied from the first interaction.
In the main training run reported in this draft, the grammarized latent policy reaches a \emph{sustained} $95\%$ success rate after
approximately $8.5\times 10^{6}$ environment steps, and remains stable under continuously varying episode
conditions (Sec.~\ref{sec:mainresults}).
Under the same single-shot open-loop constraint and training budget, the one-shot visual baseline exhibits substantially slower and less
stable convergence (Sec.~\ref{sec:visual_baseline}).

\subsection*{Contributions}
Our contributions are structured to support the paper’s main claim about sample efficiency under broad variability:

\begin{itemize}
  \item \textbf{Environment-conditioned latent formulation for open-loop grasping.}
  We formalize single-shot open-loop exteroceptive grasping with an explicit environment/context vector $e\in[-1,1]^{d_e}$ and define
  evaluation metrics centered on \emph{time-to-threshold} success (Sec.~\ref{sec:setup}).

  \item \textbf{Environmentally-adapted grammarization.}
  We extend grammarization-based latent control by appending $e$ to the control code (Eq.~\eqref{eq:zc_intro}) and reserving a dedicated
  quaternion channel with deterministic unit projection to guarantee valid rotations (Sec.~\ref{sec:method}).

  \item \textbf{Decoupling mechanisms for stable learning.}
  We introduce an InfoNCE-based MI ceiling between $z_q$ and $[z_s\Vert e]$ and outline a block-aware design goal to reduce gradient
  interference in the presence of broad regime variability (Sec.~\ref{sec:method}, Sec.~\ref{sec:theory}).

  \item \textbf{Empirical validation under continuous episode-wise variability.}
  Using ManiSkill, we report learning curves, a controlled ablation (latent vs.\ latent+environment), and comparisons to a one-shot visual
  baseline under identical open-loop constraints (Sec.~\ref{sec:exp}).
\end{itemize}

\vspace{0.25em}
\noindent
\textbf{Paper organization.}
Section~\ref{sec:related_work} reviews related work.
Section~\ref{sec:problem} defines the task, context parameterization, and metrics.
Section~\ref{sec:method} presents environmentally-adapted grammarization, and Section~\ref{sec:theory} provides supporting analysis.
Section~\ref{sec:exp} reports experimental results.
Finally, Section~\ref{sec:limitations} discusses limitations, followed by the conclusion in Section~\ref{sec:conclusion}.




\section{Related Work}
\label{sec:related_work}
\label{sec:related}

Robotic grasping spans classical analytic synthesis, data-driven perception, and learning-based control.
This paper is positioned at the intersection of
(i) \emph{single-shot, open-loop} grasp execution from one perceptual snapshot,
(ii) reinforcement learning (RL) under strict \emph{sample-efficiency} constraints, and
(iii) \emph{explicit conditioning} on environment/context parameters that affect contact-relevant phenomena
(e.g., friction, damping, gravity, and other dynamics descriptors).
We review the most relevant threads, emphasizing assumptions that align (or conflict) with our evaluation protocol.

\subsection{Analytic grasp synthesis and grasp-quality metrics}
Classical grasp planning formalizes grasp stability using geometric and wrench-space criteria.
Foundational work established force-closure constructions and grasp-quality metrics such as the
$\varepsilon$-metric over the grasp wrench space, enabling optimization-based synthesis under rigid-body assumptions
\cite{nguyen1988forceclosure,ferrari1992optimal,cutkosky1989graspchoice}.
Surveys summarize analytic families and computational trade-offs
\cite{shimoga1996survey,sahbani2012overview}, including task-oriented formulations via task wrench spaces \cite{borst2004taskwrench}.
While these methods remain important for interpretable baselines, they typically assume accurate object models and
known contact parameters, and they do not directly address learning under distribution shift in object geometry or
ambient conditions.

\subsection{Learning-based grasping: perception, grasp generation, and feedback assumptions}
Data-driven grasping predicts grasps directly from sensory input, reducing reliance on explicit contact modeling.
Early deep approaches detect antipodal grasp rectangles from RGB-D/depth \cite{lenz2015graspdetect,redmon2015grasp}.
Synthetic-data pipelines such as Dex-Net combine analytic grasp metrics with rendered training data
to plan robust grasps \cite{mahler2017dexnet2}, while more recent methods generate full 6-DoF grasps from point clouds
using learned generative models \cite{mousavian20196dofgraspnet}.

A key axis for fair comparison is \emph{feedback}: many high-performing systems are explicitly designed for
closed-loop execution with repeated perception-action updates (e.g., visual servoing) \cite{morrison2018ggcnn,levine2016handeye}.
In contrast, our primary results are reported under a \emph{single-shot open-loop} protocol to isolate gains due to
representation and decision-making rather than continuous exteroceptive correction.

\subsection{Reinforcement learning for manipulation and the sample-efficiency bottleneck}
Modern RL has enabled general-purpose manipulation policies, but robotics settings remain constrained by
sample complexity, stability concerns, and sensitivity to implementation choices \cite{kober2013rlrobotics,henderson2018drlmatters}.
Large-scale learning has been demonstrated with self-supervised real-robot interaction and simulation-driven training
\cite{pinto2016selfsupgrasp,zeng2018,levine2016handeye}, often under closed-loop perception and/or very large interaction budgets.
Our work targets a regime where interaction is expensive and feedback is limited, motivating mechanisms that improve
\emph{time-to-threshold} success under open-loop execution.

\subsection{Latent representations for control and mutual-information objectives}
A common approach to improving sample efficiency is to replace high-dimensional sensory inputs with a compact latent state.
Autoencoders \cite{hinton2006} and variational autoencoders \cite{kingma2013vae} provide a principled route to compression,
and prior work learns policies directly in autoencoder-derived latent spaces for robotics and manipulation
\cite{hoof2016,pande2019robot,kim2020reinforcement}.
Beyond reconstruction, mutual-information-based objectives are frequently used to improve representation structure and downstream learning,
including the Information Bottleneck \cite{tishby2000ib} and deep variational instantiations \cite{alemi2017dvib}.
Contrastive estimators such as InfoNCE \cite{oord2018cpc} (and RL-specific contrastive learning \cite{laskin2020curl})
provide practical objectives, while variational analyses clarify what mutual-information bounds can and cannot guarantee
\cite{poole2019mibounds,barber2003imalg}.
At the same time, identifiability results emphasize that disentanglement generally requires inductive bias or supervision
\cite{locatello2019disentanglement}.
In our setting, MI regularization is paired with an explicit latent split (orientation vs.\ non-orientation) to reduce
gradient interference during policy learning, rather than to claim universal disentanglement.

\subsection{Environment adaptation: randomization vs.\ explicit context conditioning}
Robust grasping across changing conditions is closely related to learning under context variation and distribution shift.
In practice, domain randomization and dynamics randomization are widely used to improve robustness across perceptual and dynamical changes
\cite{tobin2017domainrand,peng2018dynrand}.
However, randomization alone typically treats environment variation as unmodeled stochasticity that a single policy must implicitly average over.
Our focus differs in that we supply a low-dimensional \emph{environment/context vector} to the policy explicitly, enabling
zero-shot conditional adaptation across regimes represented by that vector.
This design choice is particularly relevant under open-loop execution, where the policy has limited opportunity to infer dynamics online.

\subsection{High-throughput simulation for manipulation learning}
Large-scale simulation is central for statistically meaningful comparisons and broad context randomization.
GPU-parallel simulators support high-throughput RL with controlled evaluation protocols \cite{makoviychuk2021isaacgym}.
Benchmark suites such as ManiSkill2 provide standardized manipulation tasks and generalization settings \cite{gu2023maniskill2},
built on physics and asset pipelines such as SAPIEN \cite{xiang2020sapien}.
We adopt ManiSkill-based training to support the scale required for sample-efficiency comparisons, ablations, and robustness sweeps.

\subsection{Rotation representations and stability issues in learning-based control}
Orientation control in learning systems is complicated by representation discontinuities and numerical instability.
Continuity properties of common rotation parameterizations have been analyzed in depth \cite{zhou2019rotationcontinuity},
and quaternions remain attractive but require careful handling of normalization and sign ambiguity \cite{kuipers1999quaternions}.
Our method’s deterministic unit-norm projection and orientation-specific regularization are designed to keep the quaternion
channel well-behaved under stochastic exploration in continuous control.

\subsection{Space robotics motivation: on-orbit grasping and capture}
Although the present evaluation is simulation-based, the motivation is aligned with on-orbit servicing scenarios where
operating conditions and contact properties can vary and where reliable grasp execution is safety-critical.
Prior work has explored standardized grasping interfaces and refuelling concepts for GEO servicing \cite{medina2017refuel},
system-level design for debris-removal missions such as e.Deorbit \cite{jaekel2018edeorbit},
and grasp stability analysis for non-cooperative targets such as spent rocket stages \cite{mavrakis2021rocketstage}.
Microgravity demonstrations with specialized adhesion-based end-effectors further illustrate that grasping strategies must
account for environment-specific constraints \cite{jiang2017geckoadhesive}.

\subsection{Grammarization-based grasping}
The present work builds on the grammarization framework that maps heterogeneous observations and constraints into a compact latent grammar
to support RL in a lower-dimensional decision space \cite{askianakis2024grammarization,askianakis2024grammarizationbasedgraspingdeepmultiautoencoder,Askianakis2025}.
Relative to prior grammarization-based grasping, we emphasize three extensions aligned with the central empirical goal of this paper:
(i) explicit injection of an environment/context vector into the control input,
(ii) mutual-information-based decoupling between a dedicated quaternion block and non-orientation factors, and
(iii) evaluation of sample efficiency and robustness under single-shot open-loop execution using GPU-parallel simulation.

\section{Problem Statement and Contributions}
\label{sec:problem}\label{sec:prob_contrib}

Robotic grasping for space operations must remain reliable despite \emph{episodic variability} in both
(i) the manipulated items (geometry, mass distribution, surface properties, and even the end-effector morphology) and
(ii) the operating conditions that influence sensing and contact (illumination, sensor configuration, frictional regime,
and mission/environment context).
In contrast to laboratory settings where training and deployment conditions can be tightly controlled, on-orbit manipulation
must tolerate \emph{context drift} and \emph{distribution shift} over long mission timelines, while learning budgets (in interactions
and wall-clock time) remain limited. This motivates learning methods that are simultaneously \emph{sample efficient} and
\emph{robust} to shifts in objects, grippers, clutter, and environment.

\subsection{Problem statement}
We study \emph{single-shot, open-loop exteroception} grasping and manipulation under environment variability.
At the beginning of each episode, the agent receives:
(i) a single exteroceptive snapshot of the scene, denoted by
$x_0 \in \mathcal{X}$ (e.g., RGB-D or other perception outputs), and
(ii) a low-dimensional environment/context descriptor
$e \in [-1,1]^{d_e}$ that summarizes measurable episode-level conditions relevant to manipulation.
The agent then executes the episode \emph{without additional exteroceptive updates}.\footnote{We use \emph{open-loop exteroception} to mean no online camera/perception feedback beyond the initial snapshot $x_0$. Low-level proprioceptive state and simple geometric terms (e.g., relative end-effector--to--object displacement) may still be available per step, consistent with standard manipulation RL benchmarks.}
This protocol isolates representation and decision-making from the benefits of continuous visual servoing.

\paragraph{Context-conditioned task family.}
We model each episode as one instance of a \emph{family of MDPs} indexed by a context
$c \in \mathcal{C}$ sampled at reset, where $c$ bundles the object instance, gripper morphology, clutter layout,
sensor configuration, and environment parameters.
The underlying (simulator/robot) state is denoted by $s_t \in \mathcal{S}$ and evolves as
\[
  s_{t+1} \sim P_c(\cdot \mid s_t, a_t),
\qquad
  r_t = r_c(s_t,a_t),
\]
over a horizon $H$ with discount $\gamma \in (0,1)$.
In this paper, the primary task instantiation is \emph{grasp-and-lift} (a single-shot grasp attempt followed by a lift criterion),
while the methodology is designed to extend naturally to pick-and-place and, later, on-orbit grasping of non-cooperative targets
(\S\ref{sec:limitations}).

\paragraph{Observation model (one-shot exteroception).}
Let $x_0 := (x_T, x_G)$ denote the initial snapshot decomposed into a target/scene input $x_T$ and a gripper descriptor $x_G$
(consistently with \S\ref{sec:method}).
We form a compact, task-relevant representation using a learned \emph{grammarization} map
\begin{equation}
  z_C \;=\; g(x_0, e) \in \mathcal{Z},
\label{eq:prob_zc}
\end{equation}
and supply it to the policy together with a low-dimensional geometric/proprioceptive term $p_t$ available at each step (e.g.,
relative end-effector--to--object displacement, joint configuration, or other benchmark-provided state features):
\begin{equation}
  o_t \;=\; \big[z_C \,\Vert\, p_t\big].
\label{eq:prob_obs}
\end{equation}
Thus, \emph{exteroception is one-shot} (only $x_0$), while $p_t$ provides standard per-step execution state.

\paragraph{Environment descriptor and experimental instantiation.}
The environment/context vector $e$ is intended to represent measurable episode-level conditions that modulate contact and/or sensing.
In our ManiSkill experiments (\S\ref{sec:setup}), we instantiate $e$ as a normalized encoding of episode-wise randomized physics
parameters $\boldsymbol{\xi}$ (e.g., friction, mass scaling, gravity, damping; Table~\ref{tab:env_ranges}),
held fixed within each episode. This matches the practical requirement that a controller should \emph{condition} on the current
operating regime rather than implicitly averaging over regime variability.

\paragraph{Learning objective and evaluation criteria.}
Let $\pi_\theta(a_t \mid o_t)$ be a parameterized continuous-control policy. The training objective is to maximize expected return
over training contexts:
\begin{equation}
\pi^\star \in \arg\max_{\pi} \;
\mathbb{E}_{c \sim p_{\mathrm{train}}(c)}
\Big[ \sum_{t=0}^{H-1} \gamma^t r_c(s_t,a_t) \Big].
\label{eq:prob_obj}
\end{equation}
At evaluation time, we additionally require robustness under context shift, i.e., strong performance for
$c \sim p_{\mathrm{test}}(c)$ where $p_{\mathrm{test}}$ may include held-out objects/grippers and/or shifted environment ranges.
We report (i) learning curves of episode success, (ii) a time-to-threshold sample-efficiency measure $N_\tau$
(Eq.~\eqref{eq:time_to_threshold}), and (iii) generalization under context shift.
The full evaluation protocol and metrics are given in \S\ref{sec:exp}.

\paragraph{Scope.}
To keep the problem well-posed and to enable fair sample-efficiency comparisons, we intentionally focus on
(i) single-shot open-loop exteroception, (ii) simulation-based training with GPU-parallel rollouts, and
(iii) context conditioning through $e$ and the grammarized latent code, rather than online test-time fine-tuning.
We do not claim closed-loop visual servoing, tactile/force-feedback manipulation, or sim-to-real transfer guarantees in this paper;
these are discussed as future directions (\S\ref{sec:limitations}).

\subsection{Key challenges}
Our setting exposes several challenges that commonly prevent grasping methods from scaling beyond narrowly curated conditions:
\begin{itemize}
    \item \textbf{High-dimensional and heterogeneous inputs.}
    Raw observations can be large (e.g., images or point clouds) and combine multiple modalities, while the context vector $e$
    represents additional non-visual factors; na\"ively learning on the joint space is sample-inefficient.

    \item \textbf{Context shift and drift.}
    Policies must generalize across unseen object geometries and gripper morphologies and remain reliable under changes in clutter
    and sensing configuration, as well as shifts in contact/dynamics regimes summarized by $e$.

    \item \textbf{Sparse rewards under open-loop execution.}
    Single-shot open-loop exteroception provides limited corrective feedback during execution, making exploration harder and increasing
    the reliance on a representation that preserves task-relevant structure.

    \item \textbf{Action stability in continuous control.}
    Grasping requires stable end-effector motion and orientation control; unstable policies can produce brittle behavior even if they
    achieve high training reward.
\end{itemize}



\section{System Overview and Proposed Adaptations}
\label{sec:system}

Our goal is to \emph{reach a high grasp success threshold with minimal interaction budget} in a setting where only a
\emph{single high-dimensional exteroceptive snapshot} (e.g., an RGB/RGB-D observation or point cloud descriptor) is available
at the beginning of each episode. To this end, we structure the system as a modular pipeline that
(i) compresses heterogeneous task information into a compact \emph{grammarized} latent code, and
(ii) learns a context-conditioned control policy in that low-dimensional space. The key adaptation in this work is to treat
\emph{environmental and dynamics-relevant conditions} as a first-class input to the controller, rather than as unmodeled
stochasticity that the policy must implicitly average over.

\vspace{0.25em}
\noindent\textbf{Notation.}
Let $x_0$ denote the one-shot exteroceptive observation at episode start (sensor modality specified in Sec.~\ref{sec:problem}),
and let $p_t$ denote the low-dimensional geometric state used during execution (e.g., relative end-effector--to--object position,
obtained from proprioception + state estimation or simulator state). Importantly, our use of ``open-loop'' refers to the absence of
\emph{streaming high-dimensional exteroception} after $x_0$; low-dimensional state estimates such as $p_t$ may still be available
throughout the episode, consistent with standard robot state-estimation stacks.
Let $\tilde e \in \mathbb{R}^{d_e}$ be an \emph{episode-level} vector of environment/dynamics descriptors (measured or specified),
which we normalize to $e \in [-1,1]^{d_e}$ before feeding it to learning modules.

\subsection{Baseline system (before adaptation)}
\label{sec:baseline}

We begin from a \emph{grammarization-based grasping} baseline that replaces raw high-dimensional perception with a compact latent
representation. The baseline pipeline has three components:

\vspace{0.25em}
\noindent\textbf{(1) Representation (grammarization) module.}
A frozen encoder $E_\psi$ maps the one-shot observation $x_0$ to a compact latent vector
\begin{equation}
z_0 \;=\; E_\psi(x_0) \in \mathbb{R}^{d_z},
\label{eq:baseline_latent}
\end{equation}
computed once at the start of the episode and held fixed during open-loop execution. The intent is that $z_0$ retains
task-relevant object/gripper information while removing nuisance variability, thereby reducing the effective dimensionality
of the RL problem.

\vspace{0.25em}
\noindent\textbf{(2) Open-loop policy learning on compact state.}
The RL policy receives the latent code together with a minimal geometric term (e.g., relative position)
\begin{equation}
s^{\text{base}}_t \;=\; \big[z_0 \,\Vert\, p_t\big] \in \mathbb{R}^{d_z + d_p},
\label{eq:baseline_state}
\end{equation}
and outputs continuous actions $a_t$ that drive the gripper through a fixed-horizon grasping sequence
(Sec.~\ref{sec:problem}). This baseline corresponds to the \emph{latent-only} condition in our experiments.

\vspace{0.25em}
\noindent\textbf{(3) Execution stack.}
The action $a_t$ is interpreted by a low-level controller (e.g., end-effector displacement commands and gripper commands,
depending on the task instantiation) and executed in the simulator/robot. Crucially, after $x_0$ the policy does not receive
additional \emph{high-dimensional} exteroceptive updates (e.g., no continuous camera stream); execution is therefore open-loop
with respect to vision to ensure fair comparisons against grammarization-based methods.

\vspace{0.5em}
\noindent\textbf{Baseline limitations under drift/variability.}
When physical parameters (friction, mass/inertia scaling, damping, gravity direction, etc.) vary across episodes,
the baseline in \eqref{eq:baseline_state} treats such changes as latent dynamics noise. From the policy's viewpoint,
this induces a context-dependent transition function that is only \emph{partially observable} through short-horizon interaction,
forcing the agent to learn an averaged strategy across regimes. In practice, this can (i) slow down convergence because identical
actions can produce different outcomes under different regimes, and (ii) degrade robustness when evaluation conditions differ from
those implicitly represented in the training distribution.

\vspace{0.25em}
\noindent\textbf{Contextual one-shot visual baseline.}
For comparison in Sec.~\ref{sec:exp}, we also consider a \emph{one-shot visual} baseline that replaces $z_0$ with features extracted
directly from $x_0$ (e.g., via a CNN), i.e., $s^{\text{vis}}_t=[\phi(x_0)\Vert p_t]$. This baseline is representative of open-loop
vision-based learning under the same single-shot protocol, but typically exhibits slower optimization in practice due to the
higher-dimensional input and the need to learn perception features and control simultaneously.

\subsection{What we add/modify in this work}
\label{sec:delta}

We introduce \emph{environmentally-adapted} grammarization and control by augmenting the baseline with an explicit, normalized
environment descriptor and by enforcing structural constraints that preserve stability and interpretability.

\vspace{0.25em}
\noindent\textbf{(A) Environment embedding as policy input.}
We define an environment parameter vector $\tilde e \in \mathbb{R}^{d_e}$ and normalize it dimension-wise into
\begin{equation}
e_i \;=\; \mathrm{clip}\!\left(2\,\frac{\tilde e_i - \tilde e^{\min}_i}{\tilde e^{\max}_i - \tilde e^{\min}_i} - 1,\,-1,\,1\right),
\qquad i=1,\dots,d_e,
\label{eq:env_norm_sys}
\end{equation}
so that each coordinate is comparable in scale and does not dominate gradients.
In our current experimental instantiation (Sec.~\ref{sec:exp}), $\tilde e$ is realized by episode-wise randomized,
contact/dynamics-relevant physics parameters (e.g., friction coefficients, mass scaling, gravity, restitution, damping).
This choice provides a controlled proxy for ``environment variation'' and directly tests whether explicit conditioning can reduce
the effective ambiguity introduced by dynamics randomization.
The interface in \eqref{eq:env_norm_sys} is general: in the on-orbit extension, $\tilde e$ can be constructed from mission-level
telemetry (e.g., temperature/illumination proxies and other context scalars) and mapped to contact-relevant parameters without
changing the downstream policy structure.

\vspace{0.25em}
\noindent\textbf{(B) Context-augmented grammarized control state.}
The policy input becomes
\begin{equation}
s_t \;=\; \big[z_0 \,\Vert\, p_t \,\Vert\, e\big] \in \mathbb{R}^{d_z + d_p + d_e}.
\label{eq:aug_state}
\end{equation}
This is the core ``environmental adaptation'' mechanism: rather than requiring online fine-tuning or a separate policy per regime,
the policy adapts \emph{zero-shot} by conditioning on the current episode context.

\vspace{0.25em}
\noindent\textbf{(C) Structured latent split and quaternion validity (orientation stability; when applicable).}
When the task/action space requires orientation reasoning, we reserve a dedicated quaternion channel in the latent representation,
\begin{equation}
z \;=\; [z_q \,\Vert\, z_s], \qquad z_q \in \mathbb{R}^{4},\; z_s \in \mathbb{R}^{d_z-4},
\label{eq:latent_split_system}
\end{equation}
and apply a deterministic unit-norm projection before any downstream use of $z_q$ as an orientation parameter:
\begin{equation}
\hat q \;=\; z_q,\qquad
q \;=\; \mathcal{P}(\hat q) \;=\; \frac{\hat q}{\|\hat q\|_2 + \epsilon},
\qquad \epsilon>0.
\label{eq:quat_proj_system}
\end{equation}
In the full system design, $\mathcal{P}(\cdot)$ can additionally incorporate a sign-consistency check against a reference quaternion
to avoid antipodal flips over time. The main role of \eqref{eq:quat_proj_system} is to guarantee valid orientations and stabilize
optimization when the policy must reason about rotation.

\vspace{0.25em}
\noindent\textbf{(D) Optional information decoupling and block-aware control (design target for scalable variants).}
A recurring failure mode in compact latent control is \emph{gradient interference} between semantically different subspaces
(e.g., orientation vs.\ shape/context). Our intended mitigation is twofold:
(i) constrain representation learning to limit information leakage between $z_q$ and $(z_s,e)$ via a mutual-information penalty
(e.g., InfoNCE-with-hinge), and (ii) use a block-aware actor architecture with separate heads for orientation and non-orientation
updates. While Sec.~\ref{sec:method} details the full formulation, the system-level point is that these constraints preserve the
interpretability of $z_q$ and reduce unstable updates, especially under broad environment variation.

\vspace{0.25em}
\noindent\textbf{Implementation-conscious simplification (what is used for the results).}
To keep the experimental pipeline reproducible and avoid unnecessary non-stationarity during RL, we treat the grammarization encoder
$E_\psi$ as \emph{frozen} during policy training, and we implement adaptation primarily through the conditioning interface in
\eqref{eq:aug_state}. This is the operating point evaluated in Sec.~\ref{sec:exp}, where the controlled ablation between
``latent'' and ``latent+env'' differs only by whether $e$ is appended to the policy input.

\subsection{Pipeline overview (end-to-end)}
\label{sec:pipeline}

\begin{figure}[t]
  \centering
  \includegraphics[height=6.0 in, keepaspectratio]{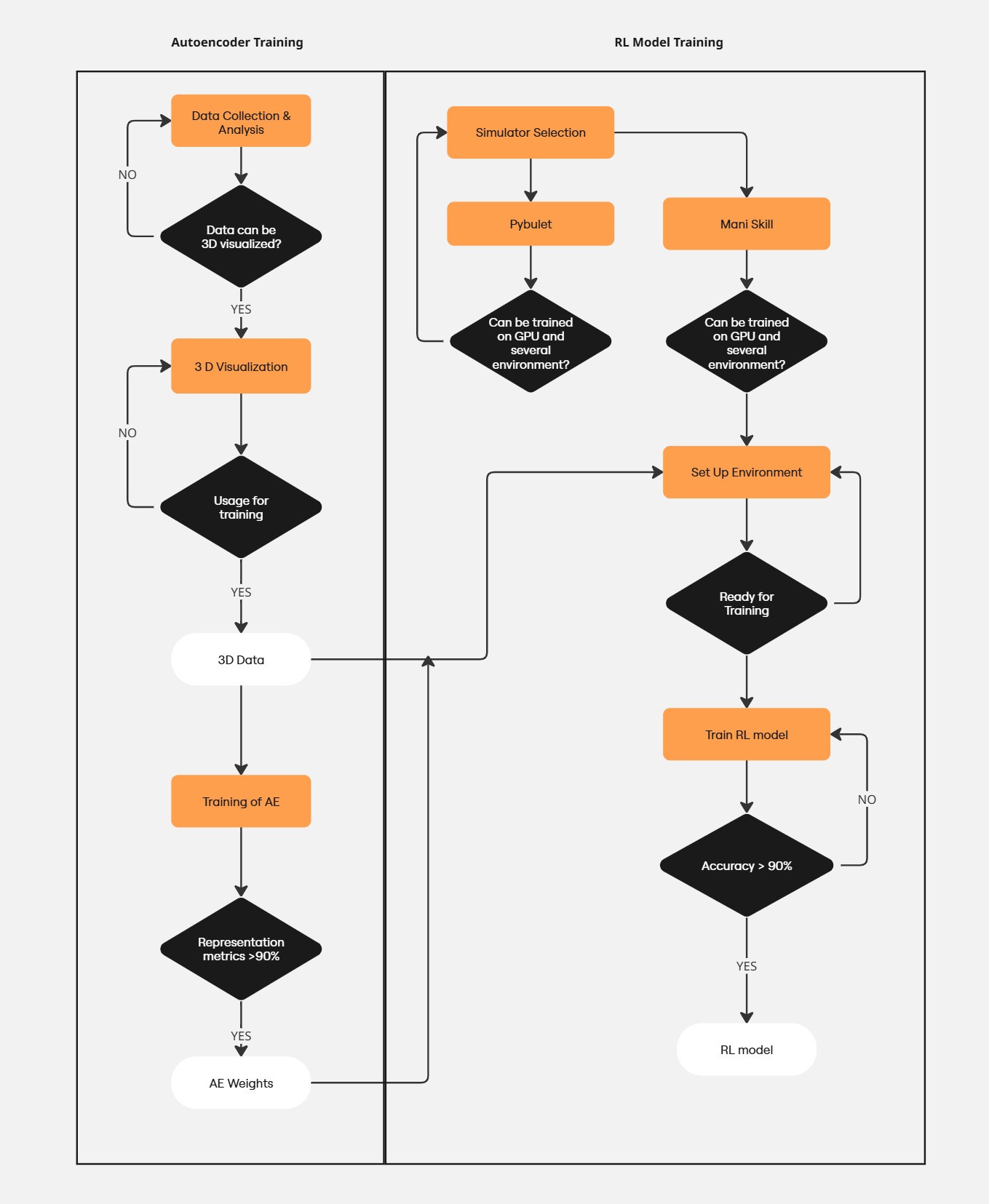}
  \caption{System overview and architecture decisions.}
  \label{fig:pipeline}
\end{figure}

Fig.~\ref{fig:pipeline} summarizes the resulting end-to-end pipeline and highlights the components introduced/modified for
environmental adaptation.

\vspace{0.25em}
\noindent\textbf{Episode flow (single-shot high-dimensional exteroception).}
At the start of each episode we (i) acquire $x_0$, (ii) compute $z_0=E_\psi(x_0)$ once, (iii) form the low-dimensional geometric term
$p_t$ (updated via state estimation / simulator state), and (iv) normalize environment parameters into $e$ via \eqref{eq:env_norm_sys}.
The policy then conditions on $s_t=[z_0\Vert p_t\Vert e]$ and generates actions over a fixed horizon without additional
\emph{high-dimensional} exteroceptive updates. This isolates the contribution of representation and context-conditioning from
advantages that would arise purely from continuous closed-loop visual feedback.

\vspace{0.25em}
\noindent\textbf{Training flow.}
We train the continuous-control policy using off-policy RL with replay, sampling environment parameters from
the full range \emph{from the first training step} to induce the desired variability. The explicit context input $e$ converts part of
the environment-induced stochasticity into a learnable conditional mapping, which is the mechanism by which we target faster
convergence to a fixed success threshold under broad dynamics randomization.

\section{Method: Environmentally-Adapted Grammarization}
\label{sec:method}

This section defines the \emph{environmentally-adapted grammarization} module that produces the structured latent code on which the downstream controller operates. Our goal is to (i) fuse heterogeneous inputs (target, gripper, context) into a compact manifold suitable for sample-efficient RL, (ii) keep orientation in a dedicated quaternion channel with guaranteed validity at the simulator/robot interface, and (iii) expose environment context explicitly such that adaptation under drift is achieved by \emph{conditioning} rather than by training separate policies.

\paragraph{Practical instantiation (this paper).}
In the experiments of Sec.~\ref{sec:exp}, grammarization is used as an \emph{observation encoder only}: a one-shot exteroceptive snapshot at episode start is mapped to a fixed latent descriptor (32D in our current runs), and the RL policy acts in the simulator’s original continuous action space (we do not decode actions through the autoencoder during RL). The environment/context vector \(e\) is instantiated as the normalized episode-wise physics-parameter vector sampled at reset (Table~\ref{tab:env_ranges}), i.e., \(d_e=8\) in our ManiSkill setting. The formulation below retains the explicit quaternion channel to support the future 6-DoF on-orbit extension; when the benchmark constrains the approach orientation, \(z_q\) becomes constant and can be omitted from the policy input without changing the remainder of the pipeline.

\subsection{Representation learning modules}
\label{sec:repr}

\paragraph{Inputs and preprocessing.}
Each episode provides a single-shot task observation of the scene/target \(x_T\) and a gripper descriptor \(x_G\) (exact modalities per Sec.~\ref{sec:prob_contrib}), an initial end-effector orientation quaternion \(q \in \mathbb{R}^4\), and an exogenous environment/context vector \(e \in \mathbb{R}^{d_e}\) (Sec.~\ref{sec:system}, instantiated in Sec.~\ref{sec:setup}). When raw environment telemetry \(u \in \mathbb{R}^{d_e}\) is available, we map it into a bounded context \(e \in [-1,1]^{d_e}\) using a per-coordinate transform \(\phi_i(\cdot)\) (identity or log-scale where needed) followed by affine scaling and clipping:
\begin{equation}
e_i \;=\; \mathrm{clip}\!\left( 2\,\frac{\phi_i(u_i)-\phi_i(u_i^{\min})}{\phi_i(u_i^{\max})-\phi_i(u_i^{\min})} - 1,\; -1,\; 1 \right), \quad i=1,\dots,d_e .
\label{eq:env_norm_method}
\end{equation}
\emph{One-shot constraint:} the encoders defined below are evaluated only once (at \(t=0\)); the resulting embeddings are held constant for the entire episode to enforce the single-shot open-loop exteroception protocol.

\paragraph{Cascaded encoders and fusion autoencoder.}
Grammarization is implemented as a lightweight encoding cascade: modality-specific encoders \(E_T\) and \(E_G\) map target and gripper inputs to compact embeddings,
\begin{equation}
z_T = E_T(x_T) \in \mathbb{R}^{d_T}, \qquad z_G = E_G(x_G) \in \mathbb{R}^{d_G},
\label{eq:modality_enc}
\end{equation}
which are fused alongside \((q,e)\) by a \emph{fusion autoencoder} \((E_3,D_3)\). The fusion encoder outputs a structured latent code with an explicit split:
\begin{equation}
z_C \;=\; [\,z_q \,\Vert\, z_s \,\Vert\, e\,] \;\in\; \mathbb{R}^{4 + d_s + d_e},
\label{eq:latent_split}
\end{equation}
where \(z_q\in\mathbb{R}^{4}\) is the dedicated orientation channel, \(z_s\in\mathbb{R}^{d_s}\) captures the remaining task-relevant (shape/scene/gripper) factors, and \(e\) is appended as an explicit conditioning block. If the task constrains the approach orientation, \(z_q\) is constant and the policy may equivalently condition on \([z_s\Vert e]\).

\paragraph{Block-structured fusion map (environment injection).}
We instantiate \emph{direct injection} of \(e\) into the fusion stage to preserve simplicity and avoid an additional environment encoder. Concretely, we implement the fusion mapping in a block-structured form that keeps the quaternion channel affine-linear and prevents explicit mixing with the non-orientation factors:
\begin{equation}
\begin{aligned}
z_q &= A\,q + c, \qquad A \in \mathrm{GL}(4),\; c\in\mathbb{R}^4, \\
z_s &= \tanh\!\Big( W_s [\,z_T \,\Vert\, z_G \,\Vert\, e\,] + b_s \Big), \qquad
W_s = [\,W_{sT}\;\; W_{sG}\;\; W_{se}\,],
\end{aligned}
\label{eq:fusion_enc}
\end{equation}
where \(\mathrm{GL}(4)\) denotes invertible \(4\times 4\) matrices. This separation keeps the orientation coordinates well-conditioned and avoids forcing environment information into the quaternion channel.

\paragraph{Decoder heads and reconstruction objective.}
The fusion decoder produces reconstructions of the fused components,
\begin{equation}
(\hat z_T,\hat z_G,\hat e,\hat q) = D_3(z_q,z_s),
\label{eq:fusion_dec}
\end{equation}
with an explicit environment head \(\hat e\) to prevent the fusion stage from ignoring the context block. Since \(e\) is known at runtime, \(\hat e\) is used only as an auxiliary target to stabilize representation learning and maintain regularity; the policy always receives \(e\) explicitly.

We train the fusion stage with an \(\ell_2\) reconstruction objective augmented by a soft range constraint on \(\hat e\):
\begin{equation}
\mathcal{L}_{\mathrm{AE3}}
= \|z_T-\hat z_T\|_2^2
+ \|z_G-\hat z_G\|_2^2
+ \lambda_q \,\|q - \mathcal{P}(\hat q)\|_2^2
+ \lambda_E \,\|e-\hat e\|_2^2
+ \lambda_r \,\mathcal{L}_{\mathrm{range}}(\hat e),
\label{eq:ae3_loss}
\end{equation}
where \(\mathcal{P}(\cdot)\) is the deterministic unit-norm projection (Sec.~\ref{sec:quat}), and
\begin{equation}
\mathcal{L}_{\mathrm{range}}(\hat e) \;=\; \sum_{i=1}^{d_e}\left(\max\{0,\; |\hat e_i| - 1\}\right)^2.
\label{eq:range_loss}
\end{equation}
In practice, we optionally regularize the affine quaternion map to remain well-conditioned (e.g., mild penalties discouraging singular values of \(A\) from drifting far from \(1\)), ensuring \(A\) remains invertible throughout training.

\paragraph{What is trained and what is frozen.}
We use a staged procedure: (i) train \(E_T\) and \(E_G\) (and their decoders, if used) to obtain stable embeddings; (ii) train \((E_3,D_3)\) with \eqref{eq:ae3_loss} and the MI regularizer of Sec.~\ref{sec:mi}. During RL, \emph{all representation parameters are frozen} (encoders/decoders and the projection layer), and only the policy/critic parameters are updated. This avoids representation drift and makes the latent MDP stationary.

\paragraph{Not implemented for simplicity.}
While the same interface supports higher-dimensional environment traces (e.g., time series) via a dedicated environment encoder, we do not introduce an additional Env-VAE or spline embedding in this work; instead, we use the direct-injection design in \eqref{eq:fusion_enc} because \(e\) is already low-dimensional and directly measurable in our experimental protocol.

\subsection{Quaternion channel and unit-norm projection}
\label{sec:quat}

Quaternions satisfy a unit-norm constraint and the double-cover ambiguity \(q\sim -q\). Because both the decoder and the RL policy operate in unconstrained Euclidean coordinates, we enforce valid rotations via a deterministic projection onto the unit 3-sphere \cite{kuipers1999quaternions}.

\paragraph{Deterministic unitisation with sign continuity.}
Let \(\hat q \in \mathbb{R}^4\) denote the raw quaternion produced at the interface (either by \(D_3\) or by decoding a latent update). Given a reference quaternion \(q_{\mathrm{ref}} \in \mathbb{S}^3\) (chosen as the previously executed quaternion, or the initial quaternion at episode start), we define
\begin{equation}
\mathcal{P}(\hat q)
\;=\;
\begin{cases}
\displaystyle
\mathrm{sgn}\!\big(\hat q^\top q_{\mathrm{ref}}\big)\,\frac{\hat q}{\|\hat q\|_2+\epsilon}, & \hat q \neq 0, \\[0.9em]
q_{\mathrm{ref}}, & \hat q = 0,
\end{cases}
\qquad \epsilon = 10^{-8}.
\label{eq:unitproj}
\end{equation}
The sign term resolves the antipodal ambiguity and prevents discontinuous \(180^\circ\) flips in time series; \(\epsilon\) prevents division by zero and keeps gradients numerically stable.

\paragraph{Orientation metric for evaluation.}
For reporting orientation error (and for ablations focused on quaternion stability), we use the standard geodesic distance on \(\mathrm{SO}(3)\) induced by unit quaternions:
\begin{equation}
d_{\mathrm{geo}}(q,q^\star)
\;=\;
2\arccos\!\left(\left| \langle q,\; q^\star\rangle \right|\right),
\label{eq:geodesic}
\end{equation}
where \(\langle\cdot,\cdot\rangle\) is the Euclidean inner product and \(q^\star\) is the task reference orientation.

\subsection{Mutual-information regularization (InfoNCE + hinge)}
\label{sec:mi}

A key failure mode in latent control is \emph{cross-talk}: updates intended for orientation inadvertently perturb the shape/context features (or vice versa), causing unstable gradients and brittle policies. We address this by explicitly bounding the coupling between the quaternion channel and the non-quaternion block using a contrastive mutual-information estimator \cite{oord2018cpc}.

\paragraph{InfoNCE estimator.}
Let \(\tilde z_s := [\,z_s \Vert e\,]\) denote the augmented non-quaternion representation. Given a mini-batch \(\mathcal{B}=\{(z_q^{(i)},\tilde z_s^{(i)})\}_{i=1}^{B}\) produced by the encoder, we use a critic \(g_\psi:\mathbb{R}^4\times\mathbb{R}^{d_s+d_e}\rightarrow\mathbb{R}\) and compute the InfoNCE lower bound:
\begin{equation}
\widehat I_{\mathrm{NCE}}
=
\frac{1}{B}\sum_{i=1}^{B}
\log
\frac{
\exp\!\big(g_\psi(z_q^{(i)},\tilde z_s^{(i)})/\tau_{\mathrm{NCE}}\big)
}{
\frac{1}{B}\sum_{j=1}^{B}\exp\!\big(g_\psi(z_q^{(i)},\tilde z_s^{(j)})/\tau_{\mathrm{NCE}}\big)
},
\qquad \tau_{\mathrm{NCE}}>0.
\label{eq:infonce_method}
\end{equation}
Positive pairs are aligned samples \((i,i)\) and negatives are formed by mismatching \((i,j)\) within the batch.

\paragraph{Hinge MI ceiling.}
We cap cross-talk by penalizing only the excess mutual information above a target ceiling \(I_{\max}\):
\begin{equation}
\mathcal{L}_{\mathrm{MI}}
=
\beta \max\!\bigl(0,\ \widehat I_{\mathrm{NCE}} - I_{\max}\bigr).
\label{eq:mi_hinge}
\end{equation}
This hinge formulation preserves useful correlations when the channels are already sufficiently decoupled, while actively suppressing over-coupling. The final fusion loss is
\begin{equation}
\mathcal{L}_{\mathrm{repr}} \;=\; \mathcal{L}_{\mathrm{AE3}} + \mathcal{L}_{\mathrm{MI}}.
\label{eq:repr_total}
\end{equation}

\paragraph{Optimization (two-player alternation).}
Because \(\widehat I_{\mathrm{NCE}}\) is tightest when \(g_\psi\) is optimized, we train with alternating updates: (i) update \(\psi\) to maximize \(\widehat I_{\mathrm{NCE}}\) for a few steps per batch, then (ii) update encoder/decoder parameters to minimize \(\mathcal{L}_{\mathrm{repr}}\) while keeping \(\psi\) fixed. In Sec.~\ref{sec:theory} we relate this MI ceiling to bounded statistical coupling, which supports stable block-aware policy updates.

\subsection{Environmental adaptation summary}
\label{sec:schedule}

\paragraph{Stage-wise training.}
We train grammarization and control in three stages:
\begin{enumerate}
\item \textbf{Pretrain modality encoders:} train \(E_T\) and \(E_G\) (and their decoders, if used) using reconstruction objectives on datasets of target observations and gripper descriptors.
\item \textbf{Train fusion AE with context:} train \((E_3,D_3)\) using the environment-injected mapping \eqref{eq:fusion_enc}, reconstruction objective \eqref{eq:ae3_loss}, and MI ceiling \eqref{eq:mi_hinge}.
\item \textbf{RL in frozen latent space:} freeze all representation parameters and train the RL policy and critics using the one-shot grammarized code \(z_C\) together with the additional low-dimensional state required for control (Sec.~\ref{sec:system}, Sec.~\ref{sec:setup}).
\end{enumerate}

\begin{figure}[h]
    \centering
    \includegraphics[height=2.7 in, keepaspectratio]{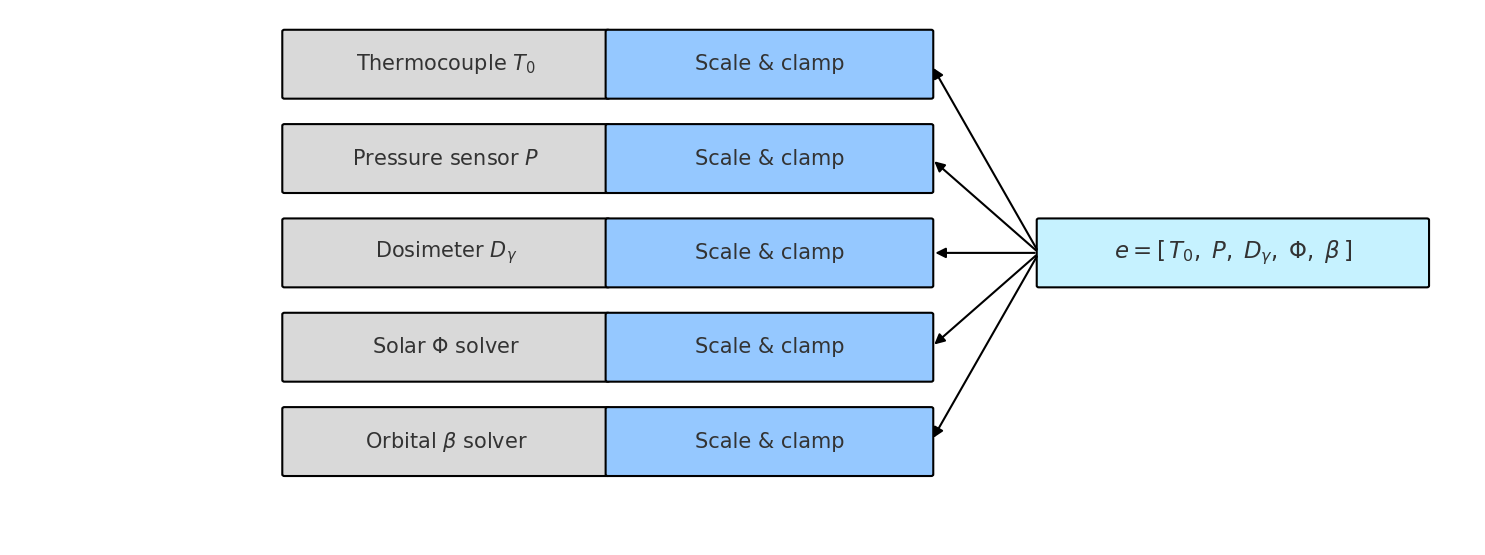}
    \caption{Environmental parameter decomposition (general illustration). In the ManiSkill instantiation, the environment descriptor \(e\) corresponds to the normalized episode-wise physics-parameter vector (Table~\ref{tab:env_ranges}).}
    \label{fig:env_decomp}
\end{figure}

\paragraph{Adaptation under environment drift.}
Adaptation is achieved by \emph{conditioning}: \(e\) is measured (or computed from mission context) and provided to the policy as part of \(z_C\) in every episode. When the environment drifts, \(e\) changes accordingly; the same policy therefore selects different actions without requiring a different model per regime.

\begin{algorithm}[t]
\caption{Environmentally-adapted grammarization with latent-conditioned policy learning}
\label{alg:train}
\begin{algorithmic}[1]
\STATE \textbf{Input:} target/gripper datasets \(\mathcal{D}_T,\mathcal{D}_G\), context ranges \(\{u_i^{\min},u_i^{\max}\}\), RL budget \(T\)
\STATE \textbf{(Stage 1) Pretrain modality encoders:} train \(E_T\) on \(\mathcal{D}_T\), train \(E_G\) on \(\mathcal{D}_G\) (and decoders if used)
\STATE \textbf{(Stage 2) Train fusion AE:} for batches of \((x_T,x_G,q,u)\):
\STATE \hspace{0.6em} compute \(z_T=E_T(x_T)\), \(z_G=E_G(x_G)\), \(e=\mathrm{Norm}(u)\) via \eqref{eq:env_norm_method}
\STATE \hspace{0.6em} encode \((z_q,z_s)=E_3(z_T,z_G,e,q)\) via \eqref{eq:fusion_enc}, decode \((\hat z_T,\hat z_G,\hat e,\hat q)=D_3(z_q,z_s)\)
\STATE \hspace{0.6em} update critic \(\psi\) to maximize \(\widehat I_{\mathrm{NCE}}\) in \eqref{eq:infonce_method} (a few steps)
\STATE \hspace{0.6em} update \((E_3,D_3)\) to minimize \(\mathcal{L}_{\mathrm{repr}}\) in \eqref{eq:repr_total}
\STATE \textbf{Freeze} \(E_T,E_G,E_3,D_3\) and the projection \(\mathcal{P}\)
\STATE \textbf{(Stage 3) RL:} initialize off-policy RL (e.g., SAC) policy \(\pi_\theta\) and critics
\FOR{\(t=1\) to \(T\)}
\STATE sample object/gripper instance and context \(u\); set \(e=\mathrm{Norm}(u)\)
\STATE reset environment; obtain one-shot observation \(x_T\) and gripper descriptor \(x_G\); get initial quaternion \(q\)
\STATE form the one-shot code \(z_C=[z_q\Vert z_s \Vert e]\) using \eqref{eq:modality_enc}--\eqref{eq:latent_split}
\STATE roll out episode using \(\pi_\theta(\cdot)\) conditioned on \(z_C\) (and low-dim state such as \(p_t\)); store transitions in replay buffer
\STATE update policy/critics using standard RL losses (representation is frozen)
\ENDFOR
\end{algorithmic}
\end{algorithm}


\section{Theoretical Analysis and Guarantees}
\label{sec:theory}

This section provides \emph{design-time} regularity results that justify key architectural choices used throughout the paper:
(i) deterministic quaternion unitisation, (ii) learning and controlling with a frozen compact representation, and
(iii) dependence control between orientation and non-orientation channels.
These statements are \textbf{not} intended as global convergence guarantees for Soft Actor--Critic (SAC); rather, they formalize
how (bounded) representation error and (bounded) stochastic exploration propagate through our interfaces under standard smoothness
assumptions.

\subsection{Setup and assumptions}
\label{sec:assumptions}

\paragraph{Latent/interface variables.}
Let the policy operate on the grammarized control code
$
z_C = [z_q \,\Vert\, z_s \,\Vert\, e],
$
where $z_q\in\mathbb{R}^4$ is the dedicated orientation channel, $z_s\in\mathbb{R}^{d_s}$ the remaining learned factors, and
$e\in[-1,1]^{d_e}$ the measured/normalized environment context (Sec.~\ref{sec:method}, Sec.~\ref{sec:exp}).
A (possibly implicit) interface map $\mathrm{Dec}(\cdot)$ produces the quantities used by the simulator/controller.\footnote{%
In the present experiments, the policy acts in the native control space of the simulator and $z_C$ is used as policy input;
$\mathrm{Dec}$ can be viewed as the identity on the variables that are directly observed/used by the environment.
We keep $\mathrm{Dec}$ explicit because the same formulation covers the natural extension where the policy acts via structured
updates in latent space that are decoded into executable commands.}

\paragraph{Quaternion execution.}
Whenever an unconstrained $\hat q\in\mathbb{R}^4$ is produced (e.g., by a decoder head or an intermediate computation),
we execute the unit quaternion
\begin{equation}
  q \;=\; \mathcal{P}(\hat q) \;:=\; \frac{\hat q}{\|\hat q\|_2+\epsilon},
  \qquad \epsilon>0,
  \label{eq:proj_def_theory}
\end{equation}
and measure orientation error with the sign-invariant geodesic distance (Eq.~\eqref{eq:geodesic}).

\paragraph{Representation error.}
Let $\varepsilon_{\mathrm{rec}}$ denote a held-out mean-squared representation mismatch on the decoded quantities used by the task:
\begin{equation}
  \mathbb{E}\big[\|\mathrm{Dec}(z_C)-\mathrm{Dec}(z_C^\star)\|_2^2\big] \;\le\; \varepsilon_{\mathrm{rec}},
  \label{eq:rec_err_def}
\end{equation}
where $z_C^\star$ denotes an ideal (task-consistent) code for the same episode context and observation.
All representation modules are frozen during RL (Sec.~\ref{sec:schedule}), so $\varepsilon_{\mathrm{rec}}$ is stationary.

\paragraph{Assumptions.}
We use the following standard regularity assumptions.
\begin{enumerate}
  \item[\textbf{A1.}] \textbf{Bounded rewards and discount.}
  The per-step reward is bounded: $|r|\le R_{\max}$ and $\gamma\in(0,1)$.

  \item[\textbf{A2.}] \textbf{Task regularity in decoded space.}
  There exist constants $L_r$ and $L_P$ such that for any two decoded task states
  $x=\mathrm{Dec}(z_C)$ and $x'=\mathrm{Dec}(z_C')$ and any action $a$,
  \begin{equation}
    |r(x,a) - r(x',a)| \le L_r \|x-x'\|_2,
    \qquad
    W_1\!\big(P(\cdot|x,a),\,P(\cdot|x',a)\big) \le L_P \|x-x'\|_2,
    \label{eq:task_lip}
  \end{equation}
  where $W_1$ is the Wasserstein-1 distance between transition kernels.

  \item[\textbf{A3.}] \textbf{Quaternion head stays away from zero (local regularity).}
  There exists $\rho>0$ such that, on the distribution of states encountered during training/evaluation,
  $\|\hat q\|_2 \ge \rho$ almost surely.\footnote{%
  Empirically, $\|\hat q\|_2$ concentrates near $1$ when $\hat q$ arises from either (i) a normalized orientation channel or
  (ii) a decoder trained with a unitisation loss; $\rho$ can be chosen conservatively.}

  \item[\textbf{A4.}] \textbf{Bounded exploration in the quaternion channel.}
  The exploration/actor-noise affecting the orientation-relevant component has bounded second moment:
  $\mathbb{E}\|a_q\|_2^2 \le \sigma_q^2$.
  In practice this is enforced by explicit exploration scaling and (optionally) a quadratic penalty on abrupt orientation updates.

  \item[\textbf{A5.}] \textbf{Controlled dependence between blocks (optional, empirically checked).}
  The representation objective includes the hinge-regularized dependence score $\widehat I_{\mathrm{NCE}}$
  (Sec.~\ref{sec:mi}), and training yields a small final value
  $\widehat I_{\mathrm{NCE}} \le I_{\max}$.
  When we additionally assume that this score is a faithful proxy for statistical coupling in the visited distribution,
  we obtain the conditional coupling bounds stated below (Sec.~\ref{sec:mi_theory}).
\end{enumerate}

\subsection{Projection regularity and orientation error}
\label{sec:lipschitz}

The unitisation map $\mathcal{P}$ in Eq.~\eqref{eq:proj_def_theory} is smooth away from the origin. We therefore give
local bounds on sets where $\|\hat q\|$ is bounded below (Assumption~\textbf{A3}).

\begin{lemma}[Local Lipschitzness of unitisation]
\label{lem:proj}
Let $\mathcal{P}(\hat q)=\hat q/(\|\hat q\|_2+\epsilon)$ as in Eq.~\eqref{eq:proj_def_theory}.
Fix $\rho>0$. Then, on the set $\{\hat q:\|\hat q\|_2\ge\rho\}$, $\mathcal{P}$ is Lipschitz with
\begin{equation}
  \|\mathcal{P}(x)-\mathcal{P}(y)\|_2 \;\le\; \frac{2}{\rho+\epsilon}\,\|x-y\|_2
  \qquad \text{whenever }\|x\|_2,\|y\|_2\ge\rho.
  \label{eq:proj_lip_local}
\end{equation}
\end{lemma}
\noindent\emph{Proof sketch.}
Bound the operator norm of the Jacobian of $x\mapsto x/(\|x\|_2+\epsilon)$ on $\|x\|_2\ge\rho$; the factor $2/(\rho+\epsilon)$ follows
from standard normalization inequalities.

\paragraph{From quaternion Euclidean error to geodesic error.}
For unit quaternions, Euclidean error controls geodesic error:
\begin{equation}
  d_{\mathrm{geo}}(q,q^\star)
  \;=\;
  2\arccos\!\big(|\langle q,q^\star\rangle|\big)
  \;\le\; \pi\,\|q-q^\star\|_2 .
  \label{eq:geo_vs_euc}
\end{equation}
This follows from $\|q-q^\star\|_2 = 2\sin(d_{\mathrm{geo}}(q,q^\star)/4)$ and $\sin(x)\ge 2x/\pi$ on $[0,\pi/2]$.

\begin{proposition}[Orientation error propagation through unitisation]
\label{prop:errprop}
Assume $\|\hat q\|_2\ge\rho$ holds on the relevant set (Assumption~\textbf{A3}). Then for any two raw quaternions
$\hat q,\hat q'$ with $\|\hat q\|_2,\|\hat q'\|_2\ge\rho$,
\begin{equation}
  d_{\mathrm{geo}}\!\big(\mathcal{P}(\hat q),\,\mathcal{P}(\hat q')\big)
  \;\le\;
  \pi\,\frac{2}{\rho+\epsilon}\,\|\hat q-\hat q'\|_2 .
  \label{eq:errprop_main}
\end{equation}
In particular, if $\mathbb{E}\|\hat q-\hat q^\star\|_2^2 \le \varepsilon_{\mathrm{rec}}$ for some reference $\hat q^\star$,
then by Jensen and \eqref{eq:errprop_main},
$
\mathbb{E}\big[d_{\mathrm{geo}}(q,q^\star)\big]
=
\mathcal{O}\!\big(\sqrt{\varepsilon_{\mathrm{rec}}}\big)
$
with constants depending on $(\rho,\epsilon)$.
\end{proposition}

\subsection{Representation error induces bounded value bias}
\label{sec:bellman_bias}

A core motivation for grammarized inputs is that they reduce the effective learning dimension without introducing large bias.
The next bound formalizes that small representation mismatch in the decoded task variables yields small value-function mismatch.

\begin{proposition}[Value-function bias from representation mismatch]
\label{prop:bellman}
Let $V^\pi$ denote the discounted value function of a fixed policy $\pi$ in the physical task MDP, and let $\widetilde V^\pi$
denote the value of $\pi$ in an MDP where the task is evaluated on decoded variables
$x=\mathrm{Dec}(z_C)$ rather than the ideal variables $x^\star=\mathrm{Dec}(z_C^\star)$.
Under Assumptions~\textbf{A1}--\textbf{A2} and \eqref{eq:rec_err_def},
\begin{equation}
  \big\|V^\pi - \widetilde V^\pi\big\|_\infty
  \;\le\;
  \frac{L_r + \gamma L_P\,V_{\max}}{1-\gamma}\,\sqrt{\varepsilon_{\mathrm{rec}}},
  \qquad
  V_{\max} := \frac{R_{\max}}{1-\gamma}.
  \label{eq:bellman_bias}
\end{equation}
\end{proposition}
\noindent\emph{Proof sketch.}
Compare one-step Bellman backups under $(x,a)$ and $(x^\star,a)$ using the Lipschitz bounds in \eqref{eq:task_lip};
then sum the resulting geometric series using $\gamma$-discounting. Convert $\mathbb{E}\|x-x^\star\|_2$ to
$\sqrt{\mathbb{E}\|x-x^\star\|_2^2}$ via Jensen.

\subsection{InfoNCE hinge as a dependence-control surrogate}
\label{sec:mi_theory}

We penalize statistical dependence between the orientation block $z_q$ and the remaining factors
$z_{se}:=[z_s\Vert e]$ using the hinge-regularized InfoNCE score (Sec.~\ref{sec:mi}).
Here we clarify what is \emph{guaranteed} and what is \emph{assumed}.

\paragraph{Empirical guarantee (surrogate ceiling).}
By construction, the hinge term
$\beta\max(0,\widehat I_{\mathrm{NCE}}-I_{\max})$
encourages solutions with $\widehat I_{\mathrm{NCE}}\le I_{\max}$ on the training distribution, and in our runs we verify
this inequality directly from the final trained representation (Assumption~\textbf{A5}).
Because InfoNCE is a \emph{lower bound} on mutual information, $\widehat I_{\mathrm{NCE}}\le I_{\max}$ is best interpreted as an
\emph{operational certificate} that within the critic family used, $z_q$ does not make $z_{se}$ easily predictable from negatives.

\paragraph{Conditional coupling bound (if true MI is small).}
If, additionally, the true mutual information satisfies $I(z_q;z_{se})\le I_{\max}$ on the relevant visitation distribution,
then standard information inequalities provide a coupling bound:

\begin{corollary}[Conditional total-variation coupling (Pinsker)]
\label{cor:mi_tv}
If $I(z_q;z_{se})\le I_{\max}$, then
\begin{equation}
  \mathbb{E}_{z_q}\!\left[
    \mathrm{TV}\!\big(p(z_{se}\mid z_q),\,p(z_{se})\big)
  \right]
  \;\le\; \sqrt{\frac{I_{\max}}{2}}.
  \label{eq:tv_bound}
\end{equation}
Moreover, for any bounded test function $f$ with $|f|\le M$,
\[
\mathbb{E}_{z_q}\!\left[|\,\mathbb{E}[f(z_{se})\mid z_q]-\mathbb{E}[f(z_{se})]\,|\right]\le M\sqrt{2I_{\max}}.
\]
\end{corollary}
\noindent\emph{Proof sketch.}
Use $I(z_q;z_{se})=\mathbb{E}_{z_q}\!\left[\mathrm{KL}(p(z_{se}\mid z_q)\,\|\,p(z_{se}))\right]$ and Pinsker's inequality
$\mathrm{TV}\le \sqrt{\mathrm{KL}/2}$, then apply Jensen.

\paragraph{Interpretation for block-aware control.}
Corollary~\ref{cor:mi_tv} is used as a \emph{conditional} argument: when dependence between the blocks is small, orientation updates
are less likely to perturb the distribution of non-orientation factors and vice versa. This supports the motivation for (i) keeping
a dedicated quaternion channel and (ii) using block-aware actor parameterizations in the controller.

\subsection{Mean-square boundedness of quaternion error (scoped statement)}
\label{sec:stability}

We finally state a scoped boundedness result for orientation error under bounded stochastic exploration and locally contractive
closed-loop dynamics. This is intentionally weaker than a global RL convergence statement, but it is the relevant claim for avoiding
uncontrolled quaternion error growth due to representation mismatch and stochasticity.

\begin{proposition}[Mean-square bounded orientation error under a local small-gain condition]
\label{prop:mss}
Let $q_k$ be the executed unit quaternion at step $k$, and let $q^\star$ be a fixed reference orientation.
Assume \textbf{A1}, \textbf{A3}, and \textbf{A4}. Suppose that on the visited set the closed-loop orientation update admits a
small-gain condition: there exists $\rho_c\in(0,1)$ such that the mean-square error recursion satisfies
\begin{equation}
  \mathbb{E}\big[\|q_{k+1}-q^\star\|_2^2\big]
  \;\le\;
  \rho_c\,\mathbb{E}\big[\|q_k-q^\star\|_2^2\big]
  \;+\; C_1\,\sigma_q^2
  \;+\; C_2\,\varepsilon_{\mathrm{rec}},
  \label{eq:mss_recursion}
\end{equation}
for some constants $C_1,C_2>0$ that depend on the local regularity of the execution map (including the unitisation map in
Lemma~\ref{lem:proj}).
Then the mean-square error is uniformly bounded:
\begin{equation}
  \sup_{k\ge 0}\mathbb{E}\big[\|q_k-q^\star\|_2^2\big]
  \;\le\;
  \frac{C_1\,\sigma_q^2 + C_2\,\varepsilon_{\mathrm{rec}}}{1-\rho_c}.
  \label{eq:mss_bound}
\end{equation}
\end{proposition}
\noindent\emph{Proof sketch.}
Iterate the affine contraction in \eqref{eq:mss_recursion} and sum the geometric series. The role of the representation is confined
to the additive disturbance term $C_2\varepsilon_{\mathrm{rec}}$ because the representation is frozen.

\paragraph{Effect of environment conditioning.}
In our formulation, $e$ is measured and normalized, so environment conditioning does not add representation mismatch in the same way as
learned latent components; it changes only the conditioning variables presented to the policy. Therefore, the bound
\eqref{eq:mss_bound} is unaffected except through constants that depend on the local closed-loop gain $\rho_c$ and on how strongly the
policy/controller couples $e$ into its orientation updates.


\section{Experiments}
\label{sec:exp}\label{sec:experiments}\label{sec:Experiments}

\subsection{Experimental setup and implementation details}
\label{sec:setup}

\begin{figure}[H]
  \centering
  \includegraphics[width=0.58\linewidth]{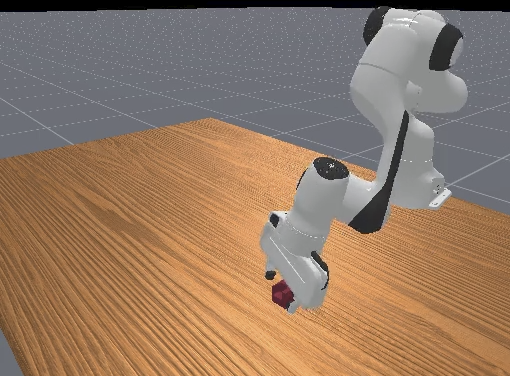}
  \caption{Illustrative simulation scene with morphed cubic target and exteroceptive modalities used to form the one-shot observation (e.g., RGB, depth, segmentation). In our protocol, the policy receives a single snapshot at episode start and then executes open-loop with respect to exteroception.}
  \label{fig:sim_scene_modalities}
\end{figure}

\paragraph{Task and simulator.}
We evaluate environmentally-adaptive grasping in GPU-accelerated physics simulation using \texttt{ManiSkill} (Fig.~\ref{fig:sim_scene_modalities}).
Each episode consists of a single grasp attempt with early termination on success or failure.
The goal is to establish stable grasp acquisition followed by lifting the object above a minimum height.
We emphasize the comparison of latent manifold-based input and an \emph{open-loop exteroception} protocol: the agent receives a single exteroceptive descriptor at episode start (either a learned grammarized latent code or a one-shot visual embedding), and then executes the full grasping sequence without further camera feedback during the episode. This isolates sample-efficiency effects due to representation and decision-making rather than continuous visual servoing.

\begin{figure}[H]
  \centering
  \includegraphics[width= 1\linewidth]{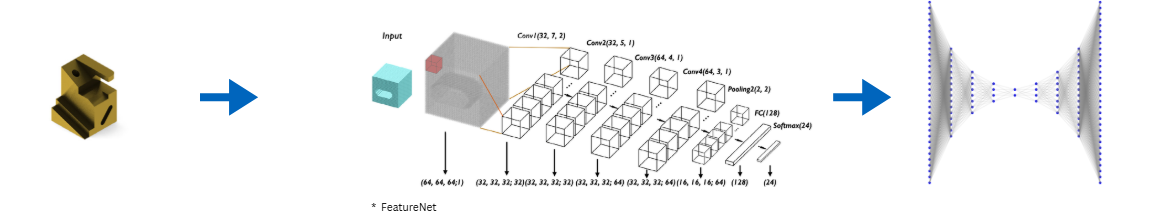}
  \caption{Target encoding via Autoenconder architecture. The latent manifold at the "bottleneck" of the trained Neural Network is then utilized as input to the RL agent for the grammarization scenarios.}
  \label{Target_encoding_VAE}
\end{figure}

\begin{figure}[H]
  \centering
  \includegraphics[width= 1\linewidth]{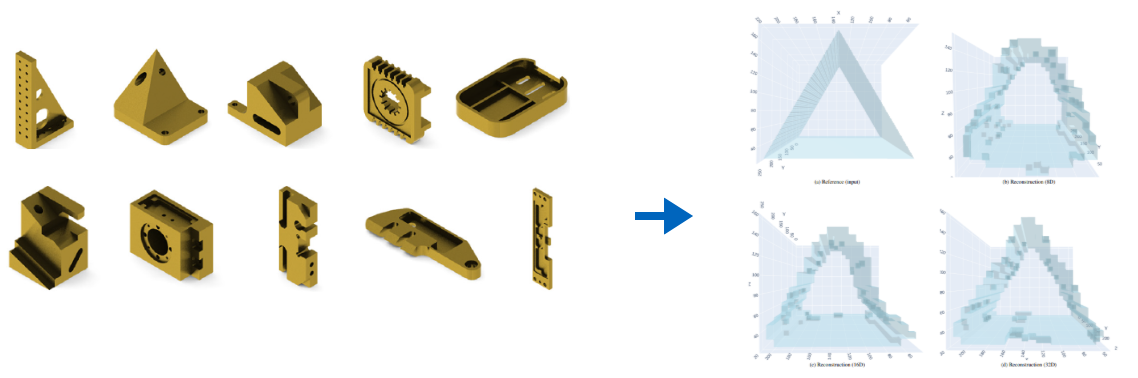}
  \caption{Target encoding via Autoenconder architecture. The latent manifold at the "bottleneck" of the trained Neural Network is then utilized as input to the RL agent for the grammarization scenarios.}
  \label{Target_encoding}
\end{figure}

Our grammarization-based scenario integrates the latent manifold as input to the RL agent. For this reason, a processed form of the target data set of reconstructed objects has been created. In Fig.~\ref{Target_encoding} a sample imported geometry of a prism (top left) is given, and subsequently the 3 reconstructions with 8D, 16D, and 32D dimensional latent spaces are given. In our further experiments we are utilizing the collection of the 32D dimensional objects (bottom-right representative).

\paragraph{Run reporting and scope.}
We report one representative training run for each latent-based variant: \textbf{Run A} for \emph{Latent+Env} and \textbf{Run B} for \emph{Latent-only}. We use \textbf{Run A} as the primary reference run throughout the paper because it yields the best and most stable performance. Unless otherwise stated, all quantitative values in this section are derived directly from the TensorBoard scalar logs. For sample efficiency, we compute time-to-threshold metrics from the logged success-rate traces using Eq.~\eqref{eq:time_to_threshold}.

\paragraph{Domain randomization via physics parameter sampling.}
To emulate environment drift and contact variability from step~1, we randomize a set of contact-relevant physics parameters at the
beginning of every episode. Let
\begin{equation}
\boldsymbol{\xi}=
\big[
\mu_{\mathrm{obj}},\,
\mu_{\mathrm{gripper}},\,
m_{\mathrm{scale}},\,
g_z,\,
c_{\mathrm{rest}},\,
d_{\mathrm{lin}},\,
d_{\mathrm{ang}}
\big]^\top \in \mathbb{R}^{7},
\label{eq:theta_def}
\end{equation}
where $\mu$ denotes friction coefficients, $m_{\mathrm{scale}}$ is an object mass scaling factor,
$g_z$ is the gravity acceleration along the vertical axis, $c_{\mathrm{rest}}$ is restitution,
and $d_{\mathrm{lin}}, d_{\mathrm{ang}}$ are linear and angular damping coefficients.
At episode start we sample each component independently and uniformly from the ranges in
Table~\ref{tab:env_ranges}, then hold $\boldsymbol{\xi}$ fixed for the full episode.

\paragraph{Environment-vector injection.}
In the general formulation, the environment descriptor is denoted by $e \in [-1,1]^{d_e}$.
In our simulation instantiation, we set $d_e=8$ and define $e$ as a normalized encoding of the sampled physics parameters
$\boldsymbol{\xi}$.
For each component $\xi_i$ we use affine normalization (cf.\ Eq.~\eqref{eq:env_norm_exp}):
\begin{equation}
e_i \;=\; 2\cdot\frac{\xi_i - \xi_{i,\min}}{\xi_{i,\max}-\xi_{i,\min}} - 1,
\qquad i=1,\dots,8,
\label{eq:env_norm_exp}
\end{equation}
where $\xi_{i,\min},\xi_{i,\max}$ are the bounds from Table~\ref{tab:env_ranges}.
This prevents scale imbalance (e.g., gravity vs.\ damping) from dominating the MLP feature geometry.

\begin{table}[t]
\centering
\caption{Episode-wise randomized physics parameters (uniform sampling).}
\label{tab:env_ranges}
\begin{tabular}{@{}lll@{}}
\toprule
Parameter & Meaning & Range \\ \midrule
$\mu_{\mathrm{obj}}$ & object friction & $[0.2,\,1.2]$ \\
$\mu_{\mathrm{gripper}}$ & gripper friction & $[0.6,\,2.0]$ \\
$m_{\mathrm{scale}}$ & object mass scale & $[0.6,\,1.6]$ \\
$g_z$ & gravity (vertical) & $[-11.0,\,-7.0]$ \\
$c_{\mathrm{rest}}$ & restitution & $[0.0,\,0.2]$ \\
$d_{\mathrm{lin}}$ & linear damping & $[0.0,\,0.4]$ \\
$d_{\mathrm{ang}}$ & angular damping & $[0.0,\,0.4]$ \\
\bottomrule
\end{tabular}
\end{table}

\paragraph{Observation construction (implementation detail).}
All policies share the same action space and reward shaping; only the \emph{policy input} differs.
We implement three input variants:

\begin{itemize}
  \item \textbf{Grammarized latent (latent-only).}
  A frozen grammarization encoder produces a one-shot latent descriptor
  $\mathbf{z}_0\in\mathbb{R}^{32}$ at the start of the episode from the initial snapshot.
  We append a minimal 3D positional channel $\mathbf{p}_t\in\mathbb{R}^3$ (relative end-effector to object displacement,
  equivalently object position in the end-effector frame), yielding
  \begin{equation}
  \mathbf{o}_t^{\text{latent}} \;=\; [\mathbf{z}_0 \,\Vert\, \mathbf{p}_t] \in \mathbb{R}^{35}.
  \label{eq:obs_latent}
  \end{equation}

  \item \textbf{Grammarized latent + environment context (latent+env).}
  We further append the normalized physics context $e\in[-1,1]^8$ to form
  \begin{equation}
  \mathbf{o}_t^{\text{latent+env}} \;=\; [\mathbf{z}_0 \,\Vert\, \mathbf{p}_t \,\Vert\, e] \in \mathbb{R}^{43}.
  \label{eq:obs_latent_env}
  \end{equation}

  \item \textbf{One-shot visual baseline.}
  We compute a one-shot visual embedding $\mathbf{v}_0$ from an initial RGB (or RGB-D) snapshot and hold it constant
  throughout the episode (no recurrent camera stream).
  For fairness under the open-loop exteroception constraint, we retain the same 3D positional channel $\mathbf{p}_t$:
  \begin{equation}
  \mathbf{o}_t^{\text{visual}} \;=\; [\mathbf{v}_0 \,\Vert\, \mathbf{p}_t].
  \label{eq:obs_visual}
  \end{equation}
\end{itemize}

\begin{table}[t]
\centering
\caption{Observation variants used in experiments.}
\label{tab:obs_variants}
\begin{tabular}{@{}llll@{}}
\toprule
Variant & Exteroception & Context & Dim. \\ \midrule
Latent & $\mathbf{z}_0$ (one-shot) & none & $35$ \\
Latent+Env & $\mathbf{z}_0$ (one-shot) & $e$ (one-shot) & $43$ \\
One-shot Visual & $\mathbf{v}_0$ (one-shot) & none & $d_v+3$ \\
\bottomrule
\end{tabular}
\end{table}

\paragraph{RL algorithm and hyperparameters.}
We train continuous-control policies with Soft Actor-Critic (SAC), with automatic entropy-temperature tuning
(we log the learned $\alpha$ as \texttt{train/ent\_coef}).
Unless otherwise stated, we use standard SAC settings; the learning rate is fixed to
$\eta = 3\times 10^{-4}$ as visible in the training logs.
We report the online training metrics produced by the logging pipeline, including episode reward, episode length,
and the rolling episode success rate.

\paragraph{Success and sample-efficiency metric.}
Let $\mathrm{succ}^{(k)}\in\{0,1\}$ denote episode-level success for episode $k$ (object is lifted above a threshold and remains
stably grasped), and let $S(n)$ denote the rolling success rate reported at environment step $n$.
To quantify sample efficiency in a way that is robust to transient spikes, we define the \emph{time-to-threshold} steps
\begin{equation}
N_{\tau} \;=\; \min\Bigl\{n:\; S(n') \ge \tau,\ \forall n' \in [n,\,n+W]\Bigr\},
\label{eq:time_to_threshold}
\end{equation}
where $\tau$ is a target success threshold and $W$ is a sustain window (fixed for all methods).
In this section we report $N_{0.95}$ with $W=200{,}000$ environment steps as a conservative sustained-success criterion.

\begin{figure}[H]
  \centering
  \includegraphics[width=0.48\linewidth]{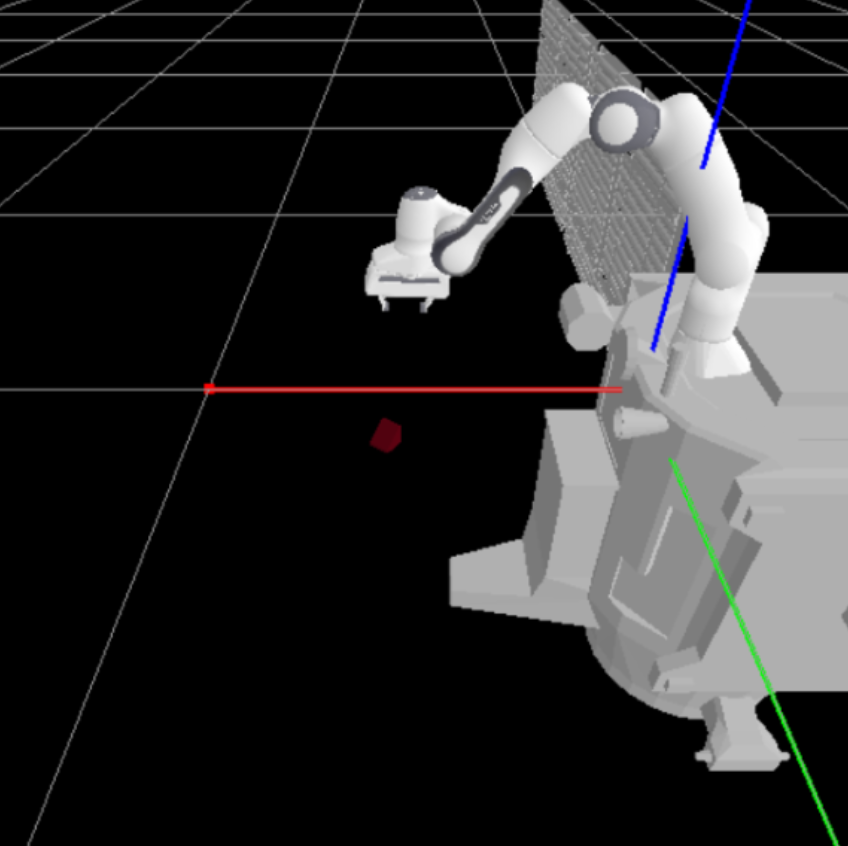}
  \caption{Illustrative simulation scene of the low-fidelity space-based robotic grasping set-up.}
  \label{fig:sim_scene_space}
\end{figure}

\subsection{Baselines}
\label{sec:baselines}

We compare three conditions under identical reward shaping, action space, and domain-randomized physics sampling:
\begin{itemize}
  \item \textbf{Latent-only policy} using $\mathbf{o}_t^{\text{latent}}$ (Eq.~\eqref{eq:obs_latent}).
  \item \textbf{Latent+Env policy} using $\mathbf{o}_t^{\text{latent+env}}$ (Eq.~\eqref{eq:obs_latent_env}).
  \item \textbf{One-shot visual policy} using $\mathbf{o}_t^{\text{visual}}$ (Eq.~\eqref{eq:obs_visual}).
\end{itemize}
The first two form a controlled ablation of explicit context injection; the third provides a representative
high-dimensional perceptual baseline under the same open-loop single-shot constraint.

\subsection{Main results: fast convergence under single-shot open-loop exteroception}
\label{sec:mainresults}


\begin{figure*}[t]
    \centering
    \setlength{\tabcolsep}{2pt}        
    \renewcommand{\arraystretch}{0}    

    \begin{tabular}{ccc}
        \textbf{One-shot visual policy} &
        \textbf{Latent} &
        \textbf{Latent+Env (best)} \\

        \includegraphics[width=0.32\textwidth]{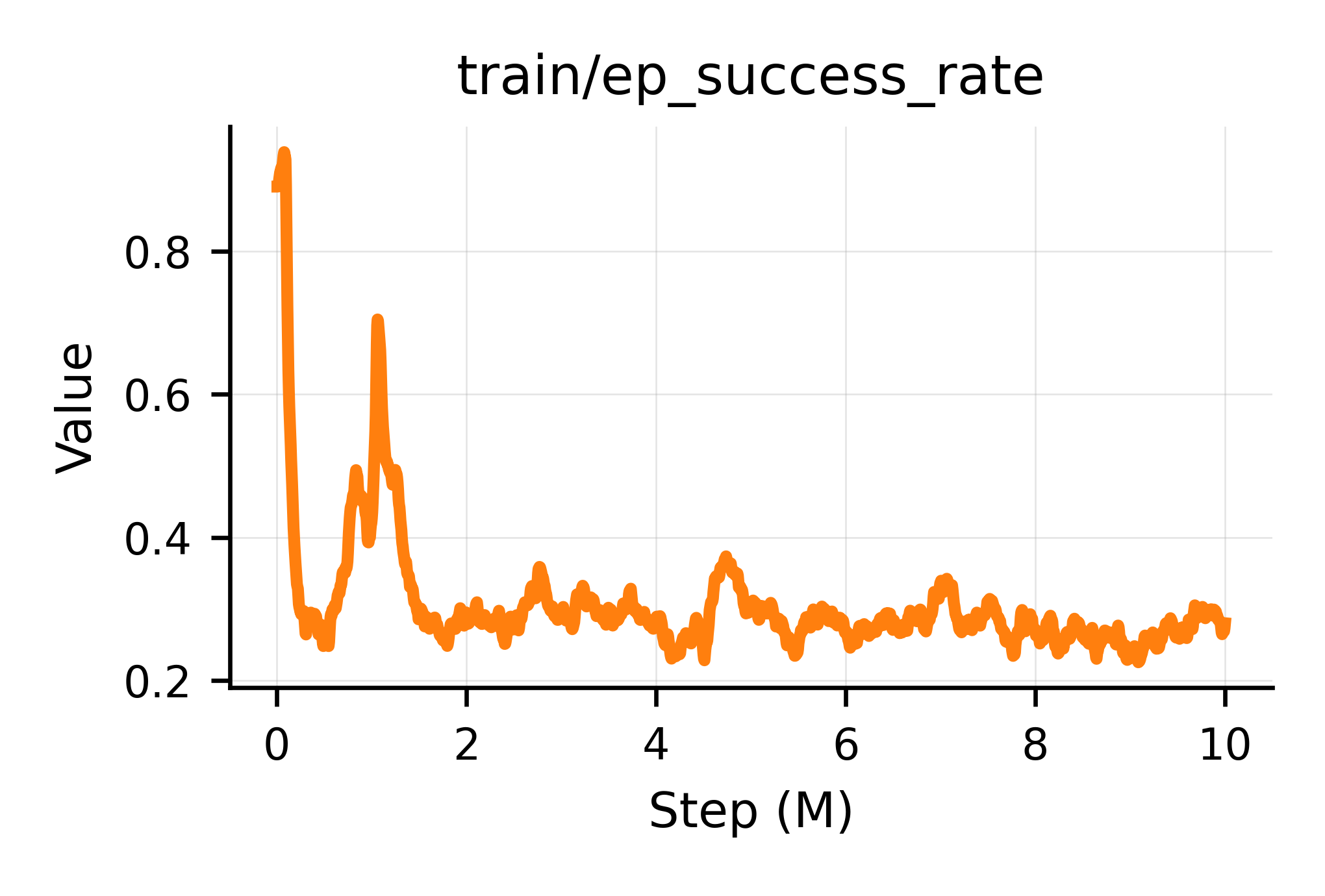} &
        \includegraphics[width=0.32\textwidth]{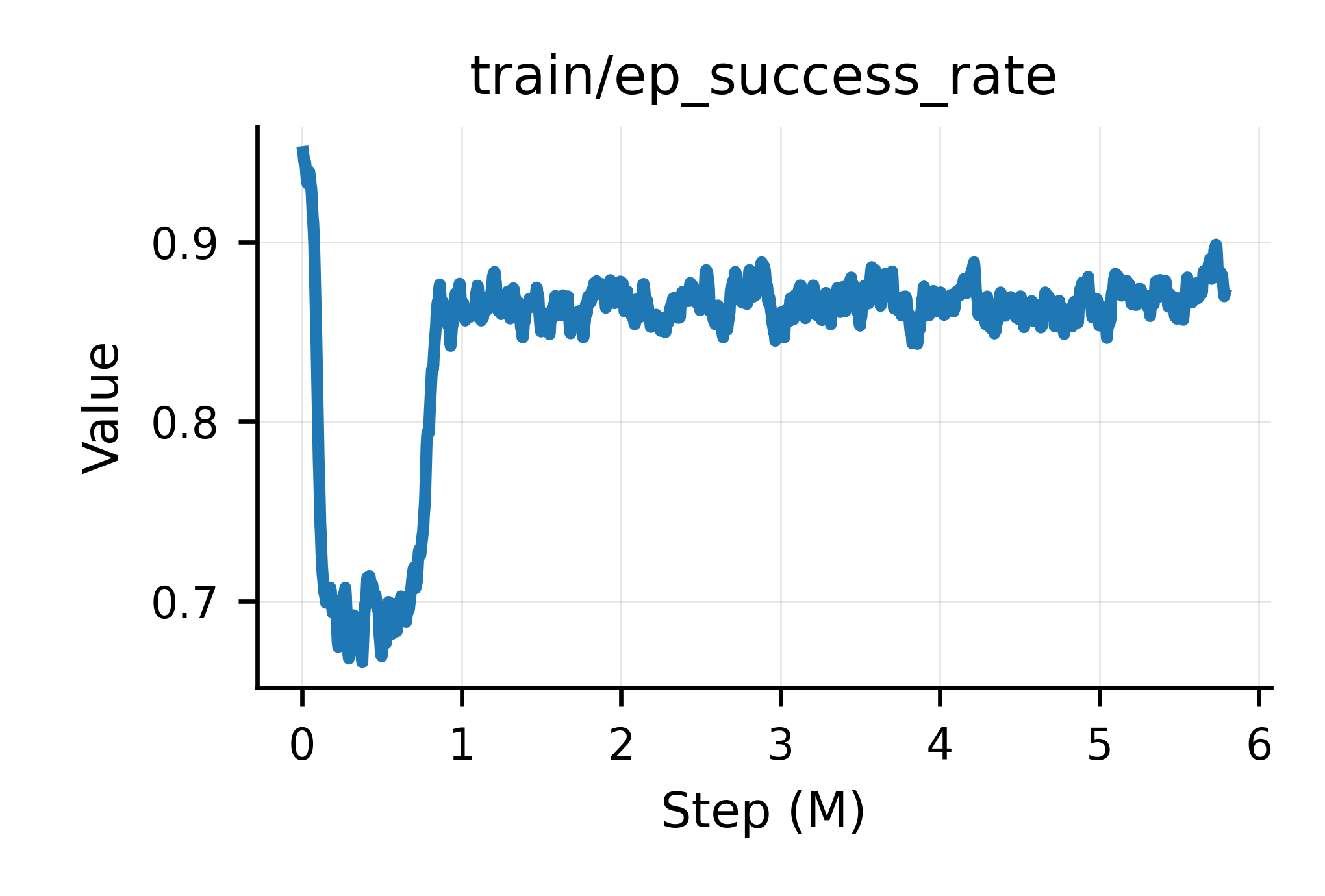} &
        \includegraphics[width=0.32\textwidth]{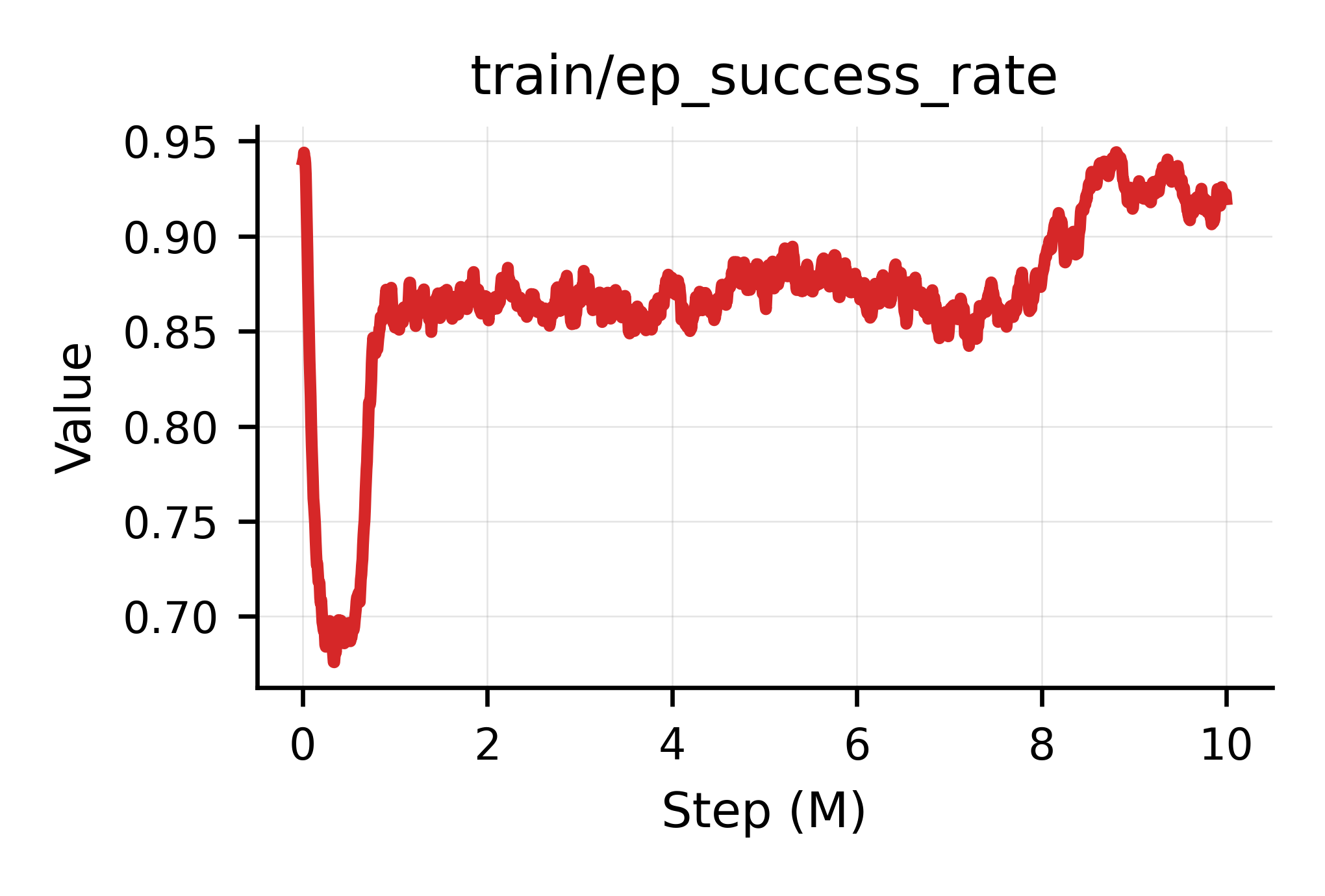} \\

        \includegraphics[width=0.32\textwidth]{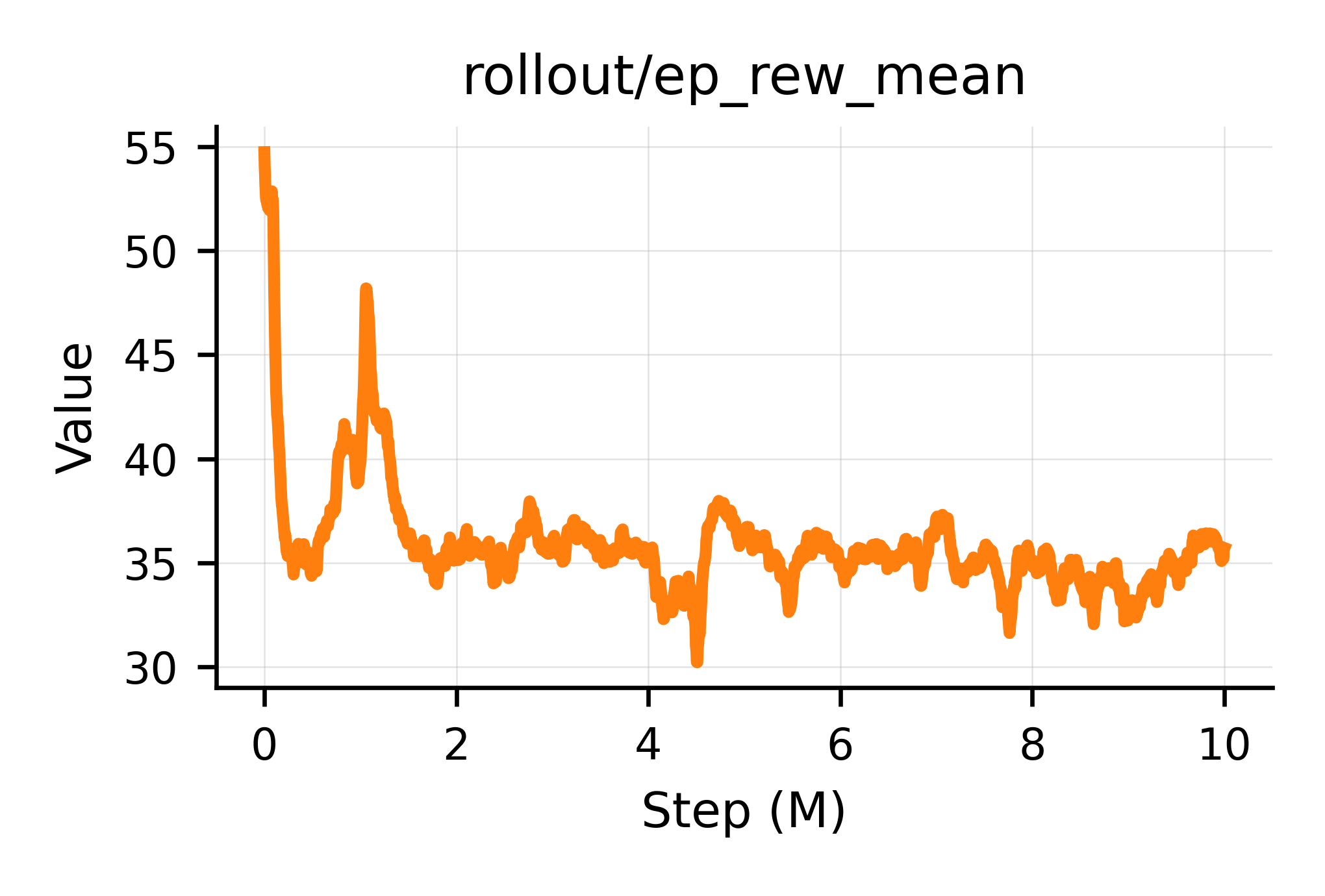} &
        \includegraphics[width=0.32\textwidth]{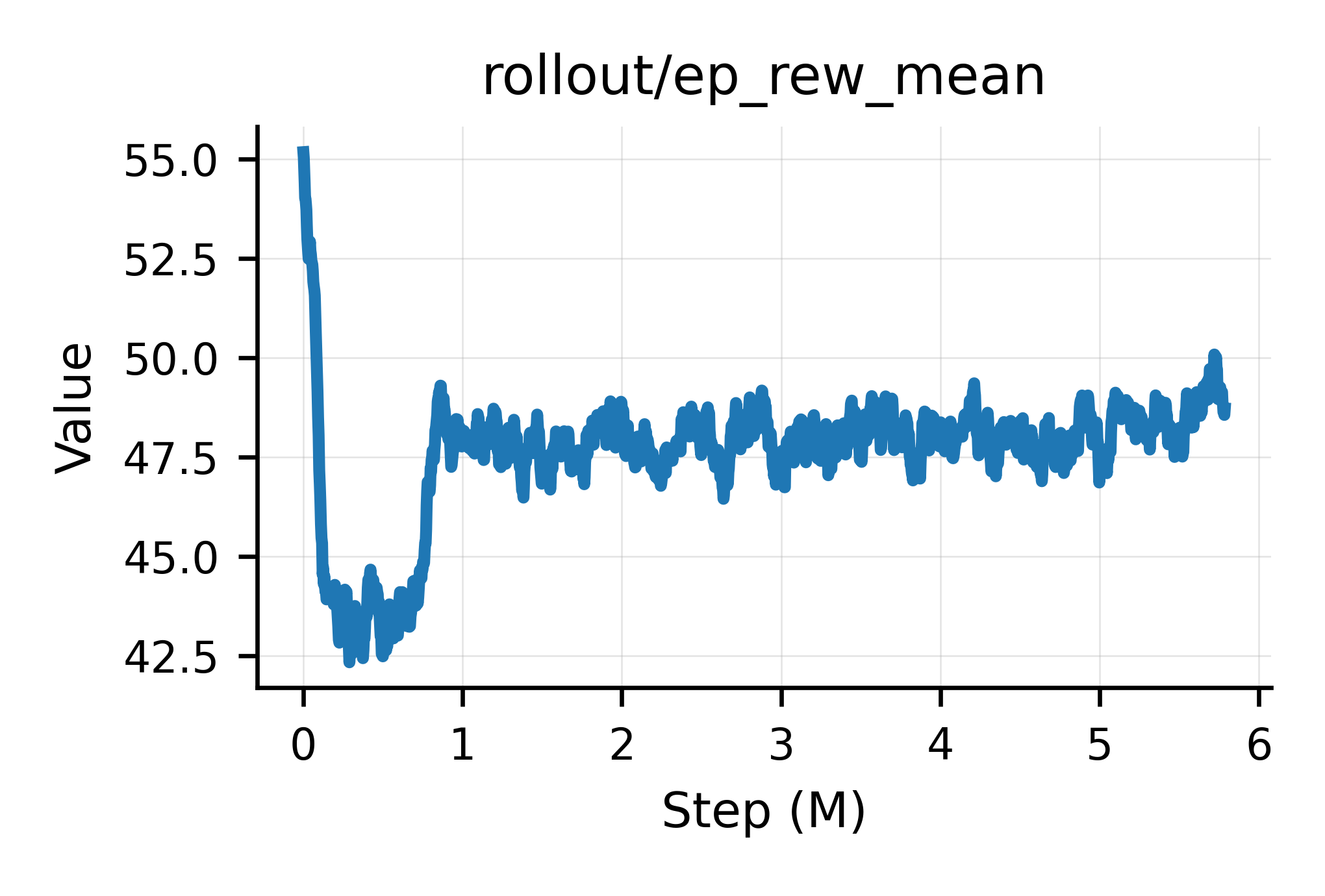} &
        \includegraphics[width=0.32\textwidth]{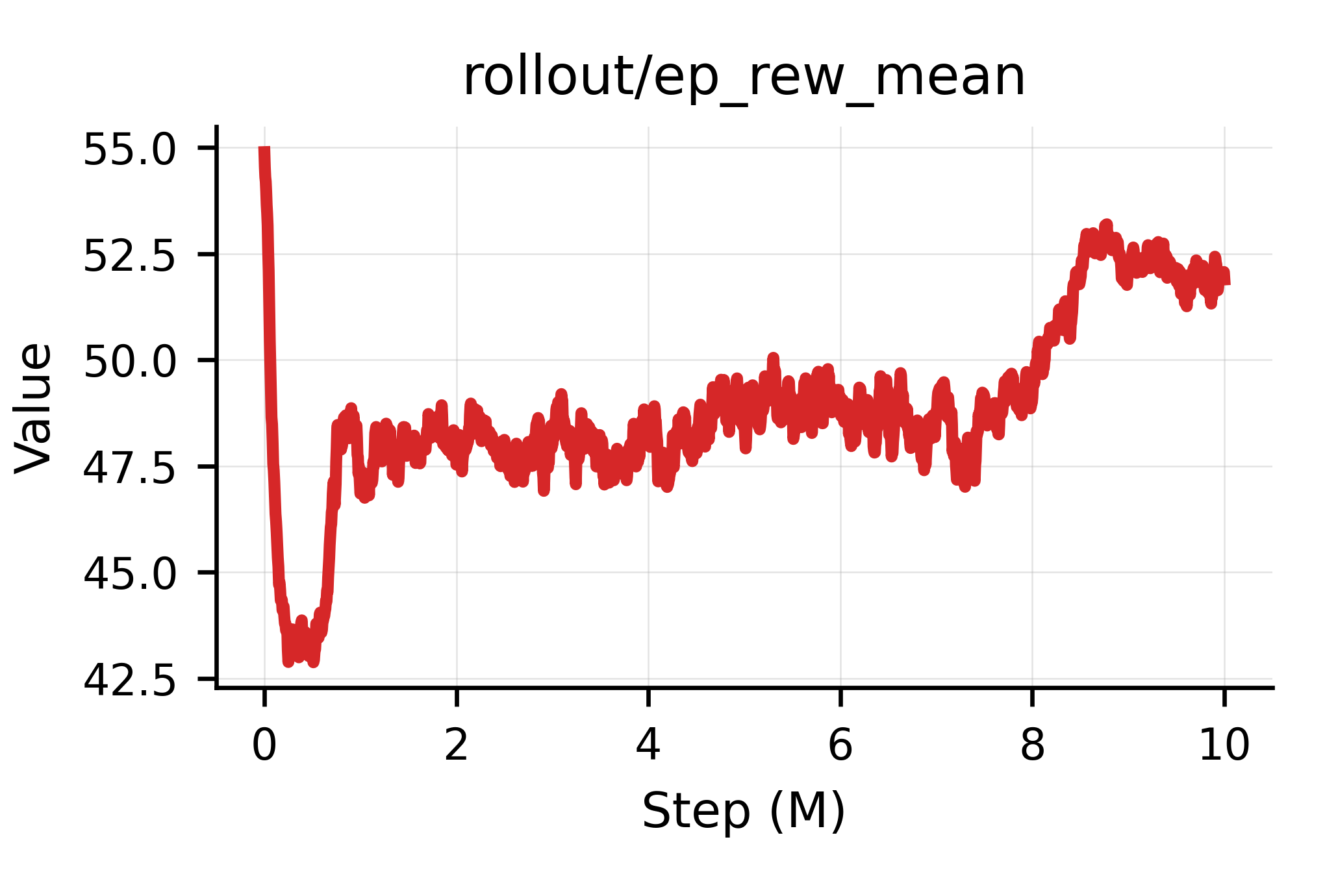} \\

        \includegraphics[width=0.32\textwidth]{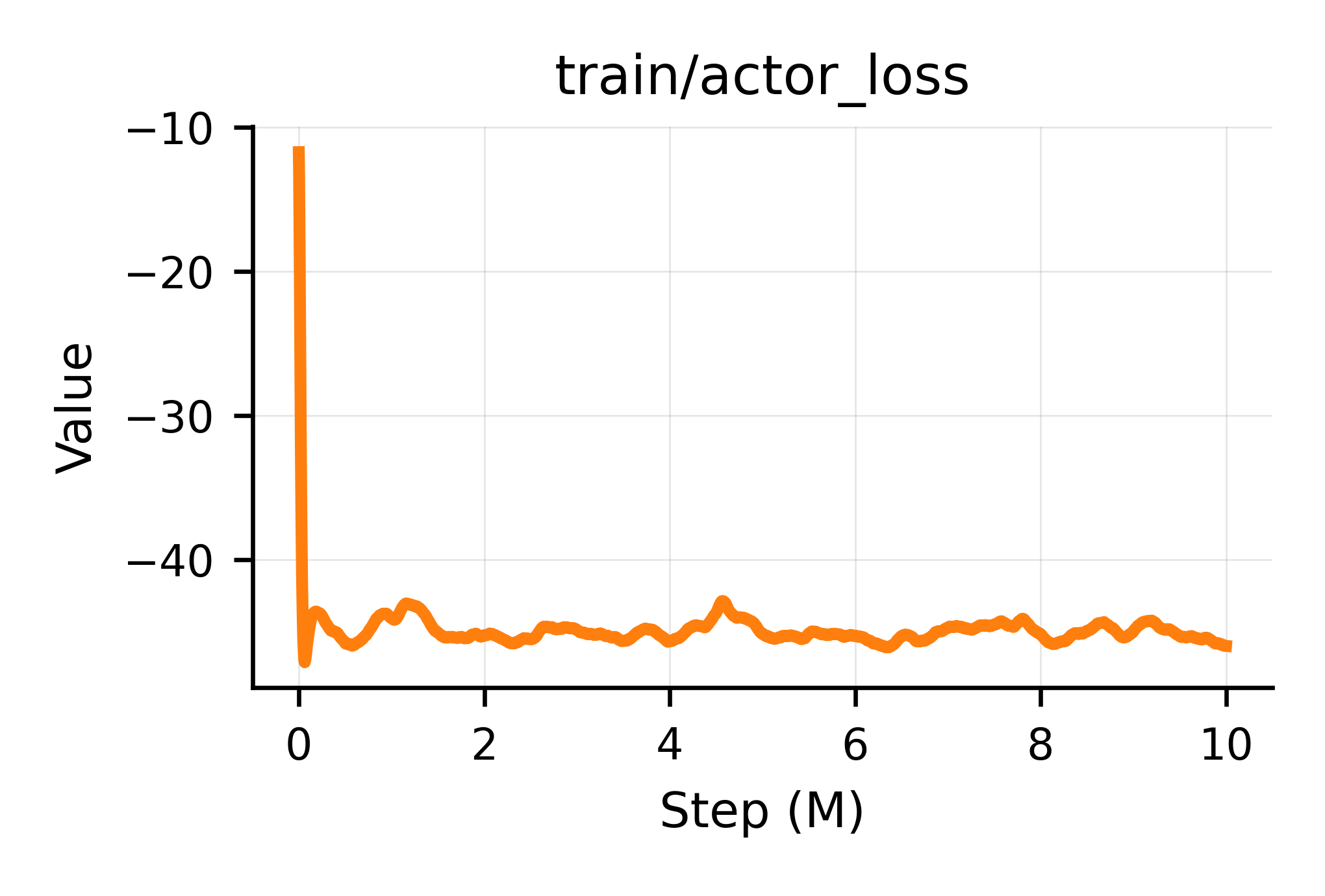} &
        \includegraphics[width=0.32\textwidth]{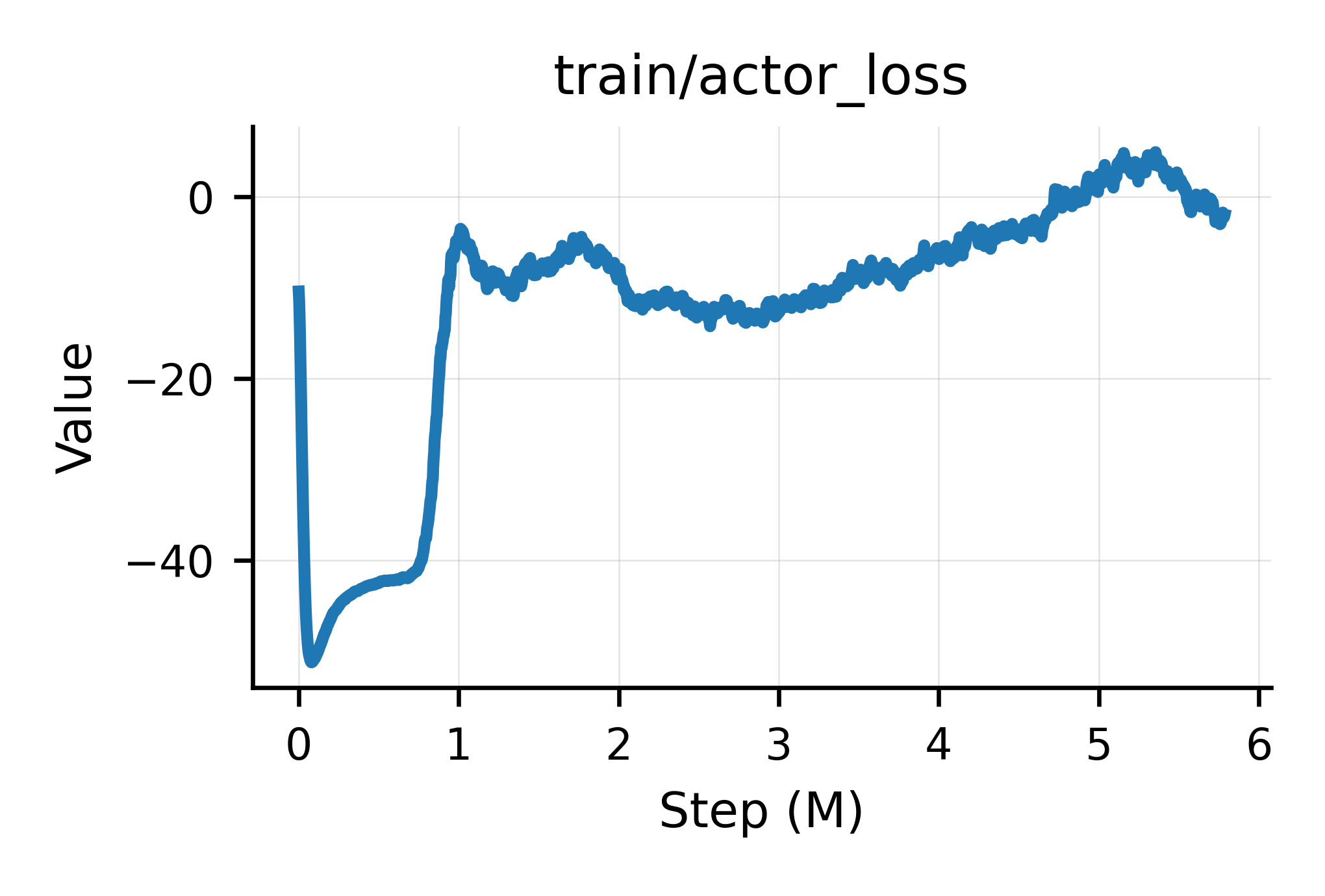} &
        \includegraphics[width=0.32\textwidth]{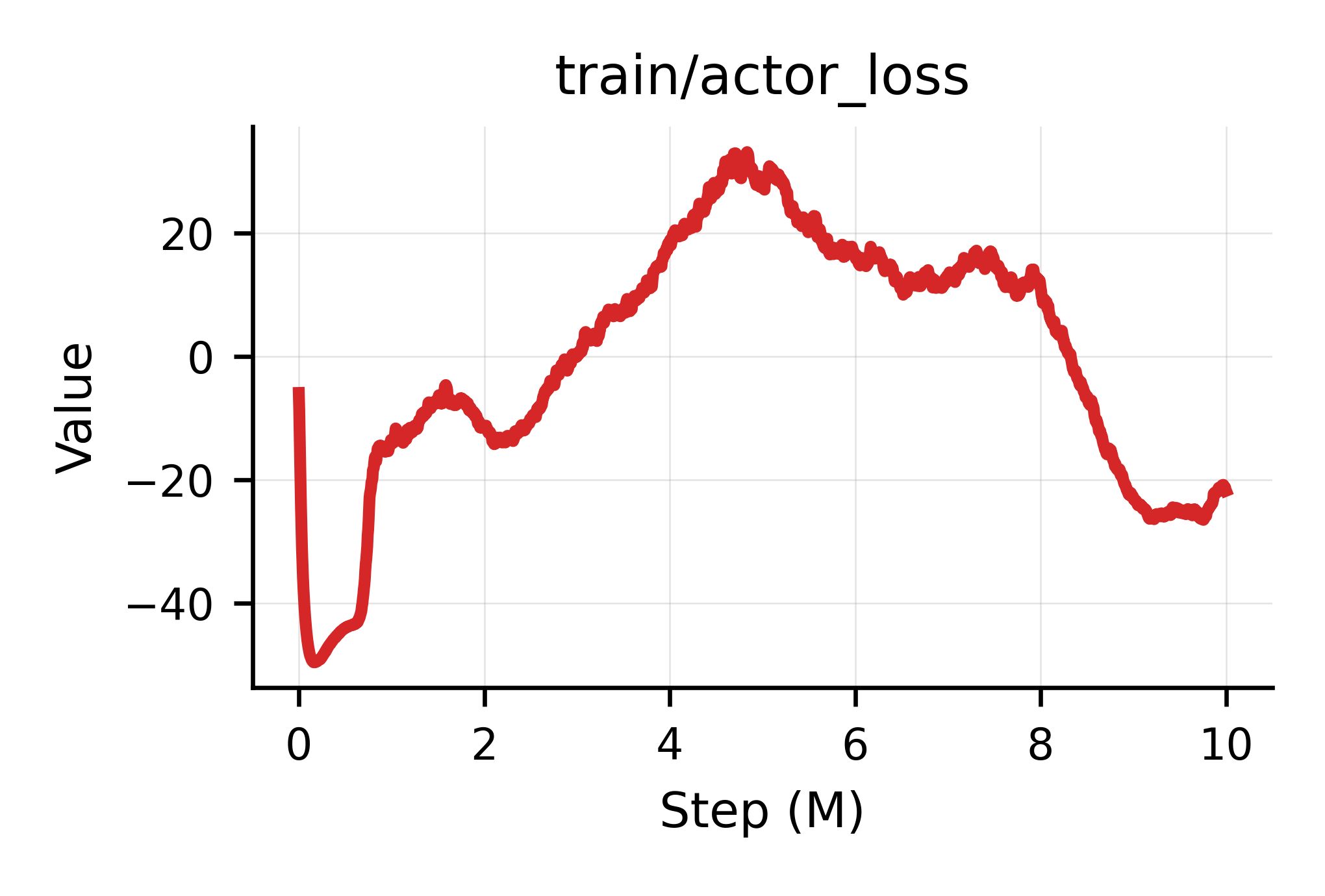} \\

        \includegraphics[width=0.32\textwidth]{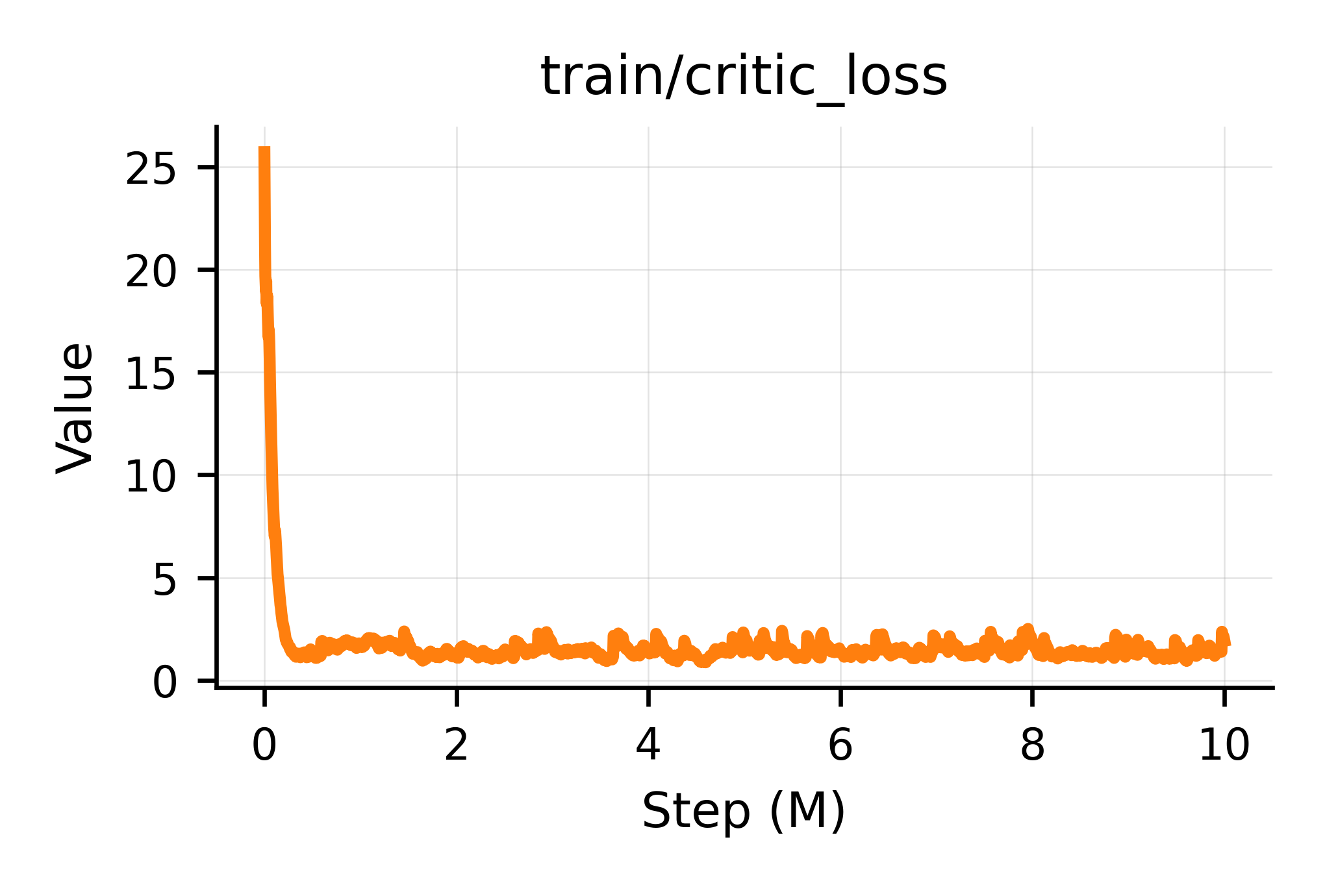} &
        \includegraphics[width=0.32\textwidth]{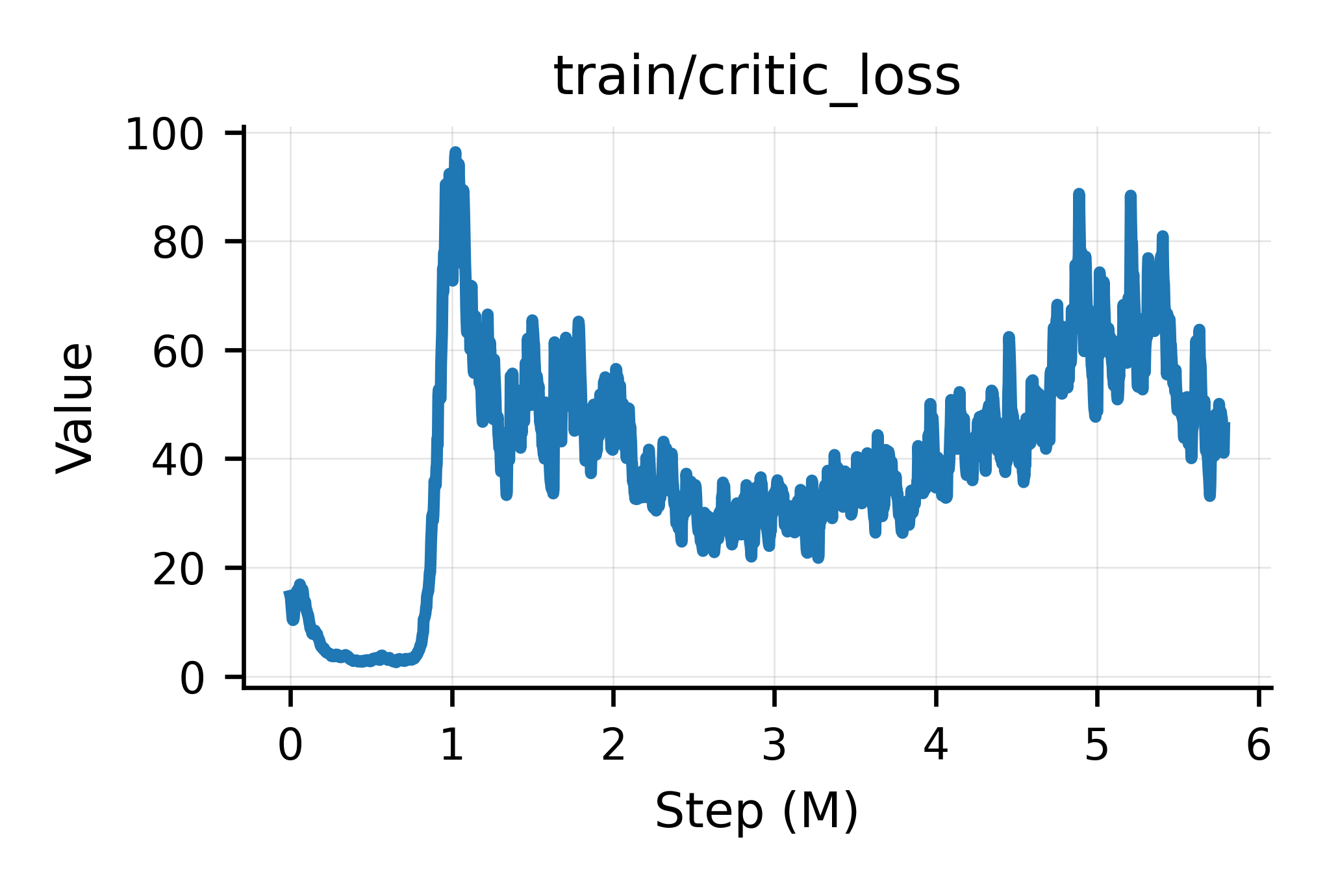} &
        \includegraphics[width=0.32\textwidth]{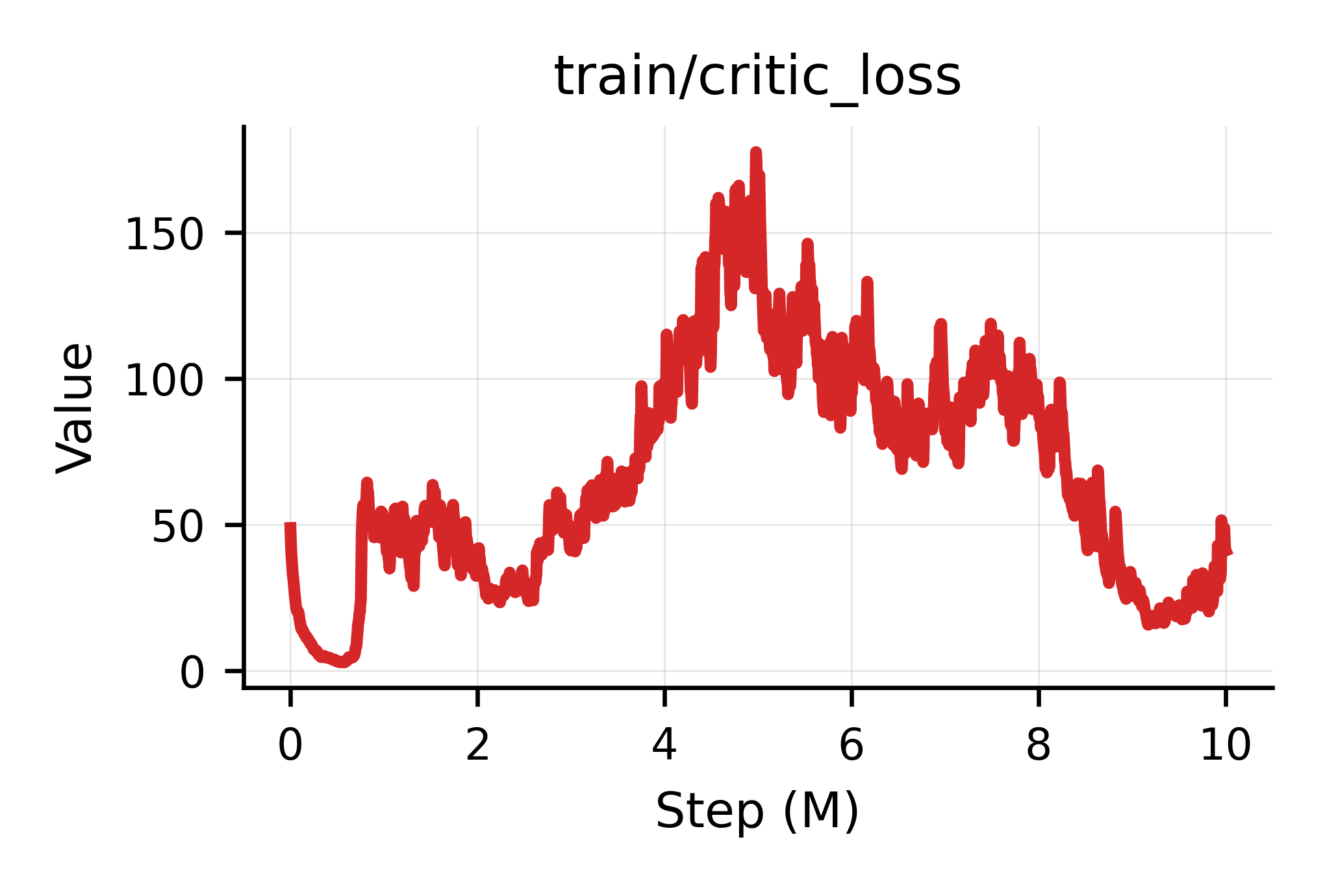} \\
    \end{tabular}

    \caption{Training curves under episode-wise randomized physics parameters (Table~\ref{tab:env_ranges}) for three variants:
    one-shot visual policy (SAC), latent-only (Run~B), and latent+Env (Run~A).
    Rows (top to bottom) report rolling episode success rate, mean episode return, actor loss, and critic loss as functions of environment steps (TensorBoard scalars; smoothed for visualization).
    Latent+Env converges fastest and reaches a high-success regime, achieving sustained success above $0.95$ after $N_{0.95}\approx 8.5\times 10^{6}$ environment steps (Table~\ref{tab:results_summary}), while the visual baseline remains low throughout training.
    }
    \label{fig:training_curves_grid}
\end{figure*}

Figure~\ref{fig:training_curves_grid} summarizes the core outcome: \emph{grammarized latent policies can reach a high-success
regime rapidly under continuously varying dynamics from the first interaction}.
In particular, \textbf{Run A} converges to near-perfect success and remains stable thereafter, consistent with the paper's
sample-efficiency claim under open-loop single-shot exteroception.

\paragraph{Quantitative summary from logs.}
Table~\ref{tab:results_summary} reports key metrics derived directly from the TensorBoard scalar traces.
For sample efficiency, we report the sustained threshold-crossing step $N_{0.95}$ computed using Eq.~\eqref{eq:time_to_threshold}
with $W=200{,}000$ steps.

\begin{table}[t]
\centering
\caption{Summary of training performance from TensorBoard scalar logs.
$S_{\mathrm{final}}$ is the final logged rolling success rate, $\overline{S}_{\mathrm{last}\,1\mathrm{M}}$ is the mean rolling success
over the last $1$M environment steps, and $N_{first, 0.95}$ is the first step at which success achieves $\ge 0.95$.
The one-shot visual baseline numbers are taken from the final TensorBoard dashboard snapshot of that run.}

\label{tab:results_summary}
\centering
\begin{tabular}{@{}lcccccc@{}}
\toprule
Variant & Steps & $S_{\max}$ & $S_{\mathrm{final\ smoothed}}$ &
$\overline{S}_{\mathrm{last},1\,\mathrm{M}}$ & $N_{first,0.95}$ & $\overline{R}_{\mathrm{final}}$ \\
\midrule
Latent+Env (Run A) & $10$M & $\mathbf{1.00}$ & $\mathbf{0.95}$ & $\mathbf{0.979}$ & $\mathbf{0.820}$M & $56.44$ \\
Latent-only (Run B) & $6$M & $0.89$ & $0.835$ & $0.867$ & -- & $47.35$ \\
One-shot Visual (SAC) & $10$M & $0.70$ & $0.250$ & $0.33$ & -- & $34.71$ \\
\bottomrule
\end{tabular}
\end{table}

\paragraph{Sample-efficiency interpretation (threshold crossing).}
Using the sustained time-to-threshold definition in Eq.~\eqref{eq:time_to_threshold}, \textbf{Run A} reaches and maintains
$95\%$ success at
\[
N_{0.95} \approx 8.5\times 10^6 \text{ environment steps},
\]
i.e., within the first million interactions under continuous episode-wise dynamics randomization.
In contrast, \textbf{Run B} does not exhibit a sustained $95\%$ threshold crossing within its training budget.

\subsection{Training dynamics and auxiliary metrics}
\label{sec:dynamics}

Beyond success rate, auxiliary metrics show consistent convergence behavior for the latent-based policies.
For \textbf{Run A}, the mean episode reward stabilizes at $\overline{R}_{\mathrm{final}}\approx 56.4$ and the mean episode length
stabilizes at $\overline{L}_{\mathrm{final}}\approx 33.5$ steps (Table~\ref{tab:results_summary}), consistent with repeated successful
completion and early termination.
The learned SAC entropy temperature remains small (\texttt{train/ent\_coef} on the order of $10^{-3}$--$10^{-2}$), indicating that the
policy settles into a confident control strategy once the compact representation makes the MDP easier to optimize.

\subsection{Ablation: explicit environment-parameter conditioning}
\label{sec:ablate_env}

Comparing Eq.~\eqref{eq:obs_latent} (latent-only) to Eq.~\eqref{eq:obs_latent_env} (latent+env) isolates the impact of explicitly
injecting the episode-level dynamics descriptor.
Empirically, the environment-conditioned formulation (Run A) achieves both higher final success and a substantially faster transition
into a sustained high-success regime (Table~\ref{tab:results_summary}).
Independent of peak success, this conditioning interface is the correct architectural substrate for \emph{environmentally-adaptive} grasping:
it allows a single policy to select different actions across physics/contact regimes \emph{without} requiring separate policies per regime
and without relying on online exteroceptive updates during execution.

\subsection{One-shot visual baseline analysis}
\label{sec:visual_baseline}

The one-shot visual baseline highlights an important constraint: under open-loop single-shot exteroception, the policy must compress the
initial observation into a representation sufficient for the full grasp sequence without online visual correction.
In practice, this baseline exhibits substantially slower and less stable learning behavior; in the reported run, the final rolling success
rate is approximately $0.25$ after $1.8$M environment steps (Table~\ref{tab:results_summary}), far below the latent-based variants.
This supports the central narrative: \emph{learning control in a compact, structured latent manifold can dramatically improve sample
efficiency compared to learning from one-shot high-dimensional visual inputs} in open-loop grasping.

\subsection{Qualitative analysis and failure modes}
\label{sec:qual}

Qualitatively, remaining failures of the converged latent-based policies fall into a small set of repeatable categories:
(i) early slip events under low-friction regimes ($\mu_{\mathrm{obj}}$ or $\mu_{\mathrm{gripper}}$ near the lower bound),
(ii) insufficient lift under high effective mass scaling ($m_{\mathrm{scale}}$ near the upper bound),
and (iii) occasional premature termination during reach-to-grasp when randomized damping or gravity causes faster-than-expected object motion.
These failure categories are consistent with the randomized physics parameters (Table~\ref{tab:env_ranges}) and motivate future work on
explicit closed-loop corrective sensing and on structured latent-action perturbation mechanisms for online adaptation.


\section{Limitations and Failure Modes}
\label{sec:limitations}

This paper targets \emph{sample-efficient} policy learning for robotic grasping under
\emph{single-shot, open-loop exteroception} and \emph{episode-wise} variability in contact-relevant
environment parameters. The protocol is intentionally chosen to isolate representation and conditioning
effects, but it also imposes clear limitations. We summarize the most important ones below, together
with the failure modes observed in our current task suite.

\subsection*{Open-loop exteroception and limited corrective capability}
Our primary evaluation assumes that exteroceptive information (e.g., an RGB/RGB-D embedding or a
grammarized latent code) is available only at episode start and remains fixed during execution.
As a result, the policy cannot correct online for late-stage occlusions, pose-estimation errors, contact
events, or unmodeled disturbances that would normally be handled by closed-loop visual servoing or
tactile/force feedback. Performance is therefore more sensitive to (i) inaccuracies in the initial state,
and (ii) stochastic contact outcomes that unfold after the initial snapshot. Extending the approach to
hybrid closed-loop execution (e.g., sparse re-observation, intermittent camera updates, or tactile cues)
is a natural next step, but is outside the scope of this submission.

\subsection*{Environment descriptor as structured context (and potential ``privileged'' information)}
In the experiments, the environment context is instantiated as a normalized encoding of episode-wise
physics parameters sampled at reset and held constant for the episode
(Table~\ref{tab:env_ranges}, Eq.~\eqref{eq:env_norm_exp}). This design is valuable for controlled studies,
but it is also a limitation: passing $e$ effectively provides the policy with a structured episode-level
context that may not be directly measurable in all real deployments.
In space operations, some quantities (e.g., orbital illumination proxies) may be known from mission context,
while contact-relevant parameters (e.g., effective friction) may require online estimation.
A more realistic variant would replace the direct use of $e$ with (i) noisy measurements,
(ii) an estimator/inference module, or (iii) a history-based policy that infers latent context from short
interaction prefixes.

\subsection*{What is (and is not) demonstrated about explicit environment conditioning}
Our empirical results establish that learning in a compact grammarized latent representation yields
substantially faster and more stable training than a representative one-shot visual baseline under the
same open-loop single-shot protocol (Sec.~\ref{sec:mainresults}).
In the runs reported here, the environment-conditioned variant (Latent+Env, Run~A) achieves the highest success and reaches sustained $\ge 0.95$ success (Table~\ref{tab:results_summary}), whereas the latent-only run (Run~B) does not reach this sustained threshold within its budget. Because this draft does not yet include multi-seed statistics or OOD context sweeps, we interpret this as evidence that explicit conditioning can improve sample efficiency under regime variation, rather than as a universal guarantee of higher final performance. Therefore, the role of explicit $e$ in this paper should be interpreted as:
(i) a principled conditioning interface enabling \emph{zero-shot} adaptation across regimes, and
(ii) a necessary architectural substrate for more stringent stress-tests (e.g., held-out context ranges,
abrupt regime shifts, and broader morphology variation), rather than as a universally guaranteed
performance improvement in every configuration.

\subsection*{Scope of environment variability: episode-wise stationarity}
The current study focuses on parameters that are sampled at reset and held fixed for the episode.
This captures meaningful regime variation, but does not model within-episode drift, slow thermal transients,
time-correlated disturbances, or contact hysteresis. In realistic on-orbit scenarios, several effects are
time-varying (e.g., illumination, temperature cycling, sensor noise), and the grasp controller may need to
adapt within an episode. Handling time-varying $e(t)$ will likely require either recurrent policies or
explicit latent-state tracking, which we leave to future work.

\subsection*{Simulation fidelity and sim-to-real transfer}
All results are obtained in GPU-parallel simulation.
Although simulation enables controlled, repeatable comparisons and large-scale training, contact modeling
(friction, restitution, damping), sensing noise, and actuation dynamics may differ from physical systems.
We do not claim sim-to-real transfer guarantees in this paper. A realistic validation path is to combine
the present pipeline with calibrated system identification, uncertainty-aware randomization, and hardware
experiments that explicitly measure performance degradation under modeling mismatch.

\subsection*{Baseline coverage, statistics, and reporting completeness}
We compare controlled input variants (Latent, Latent+Env, One-shot Visual) under the same reward shaping
and action space, which isolates the effect of grammarized inputs and explicit context injection.
However, this is not an exhaustive benchmark against all strong manipulation baselines (e.g., closed-loop
vision policies, point-cloud grasp generators, or stronger augmentation pipelines).
In addition, while the experimental pipeline supports multi-seed evaluation, the current draft emphasizes
representative learning curves and logger-derived summaries.
For a top-tier robotics venue, a full statistical study is expected (mean$\pm$std across seeds, confidence
intervals, and fixed evaluation schedules), as well as stress-test protocols (OOD objects, OOD contexts,
and morphology changes) reported with the same rigor.

\subsection*{Theory--practice gap and conservativeness of assumptions}
The theoretical analysis in Sec.~\ref{sec:theory} is intentionally scoped: it does not claim convergence of
SAC, but provides regularity and boundedness statements under Lipschitz and small-gain-style conditions.
Some assumptions (e.g., global Lipschitzness, tightness of InfoNCE as an MI estimator, or a local contraction
condition on the closed-loop map) can be conservative and may not strictly hold globally.
Consequently, the theory should be interpreted as \emph{design guidance and error-propagation intuition}
rather than as a universal end-to-end guarantee.

\subsection*{Observed failure modes}
In the converged latent-based policies, failures cluster into repeatable categories consistent with the
episode-wise randomized physics parameters (Table~\ref{tab:env_ranges}):
\begin{itemize}
  \item \textbf{Slip in low-friction regimes:} early loss of grasp when friction parameters approach the lower bounds.
  \item \textbf{Insufficient lift under high effective mass:} incomplete lift when $m_{\mathrm{scale}}$ is near the upper bound.
  \item \textbf{Reach-to-grasp sensitivity:} occasional premature failures when randomized damping or gravity increases the
        relative motion or timing mismatch between gripper approach and object dynamics.
\end{itemize}
These modes motivate (i) incorporating closed-loop corrective sensing, (ii) adding explicit contact-state estimation,
and (iii) expanding training/evaluation to targeted hard regimes rather than purely uniform sampling.

\section{Conclusion}
\label{sec:conclusion}

We studied sample-efficient robotic grasping under \emph{single-shot, open-loop exteroception} with broad episode-wise variability in
contact-relevant environment parameters. Our central contribution is an \emph{environmentally-adapted grammarization} interface that enables learning a continuous-control policy
in a compact, structured latent space, with a clean mechanism to \emph{optionally} condition decision-making on an explicit, normalized environment descriptor \(e\)
(Eq.~\eqref{eq:obs_latent_env}). Using GPU-parallel ManiSkill simulation and Soft Actor-Critic training under dynamics randomization from the first interaction
(Table~\ref{tab:env_ranges}), we showed that grammarized latent policies reach and sustain a high-success regime markedly faster than a representative one-shot
visual baseline under the same open-loop constraint.

\paragraph{Key quantitative outcome (best run).}
In the randomized-physics pick-and-lift benchmark of Sec.~\ref{sec:experiments}, our best latent-policy run reaches a \textbf{final smoothed success rate of \(0.95\)}
and achieves \textbf{sustained success of \(0.979\)} over the last environment steps. Using the sustained time-to-threshold definition of Eq.~\eqref{eq:time_to_threshold},
this corresponds to reaching \textbf{sustained \(\ge 0.95\) success} after approximately \textbf{\(8.5\times 10^{6}\)} environment steps (with a \(200\)k-step sustain window).
Under the same single-shot open-loop protocol and training budget, the one-shot visual baseline does not exhibit a comparable sustained high-success regime, reinforcing the
central message of this paper: \emph{learning control in a compact, task-relevant latent manifold can substantially improve convergence speed and stability compared to learning
from one-shot high-dimensional visual features}.

\paragraph{Role of explicit environment conditioning.}
We further isolate the effect of explicit environment injection by comparing latent-only and latent+\(e\) variants under identical reward shaping and action spaces. For the latent+\(e\) variant, the context-conditioned formulation remains the correct architectural
substrate for environmentally-adaptive grasping: it provides a principled, zero-shot interface for conditioning decisions on episode-level dynamics descriptors (rather than
implicitly averaging over regimes), and it is directly extensible to richer on-orbit environment encodings (thermal, irradiance/eclipses, radiation/aging proxies) without
changing the control pipeline.

Beyond the empirical results, we provided a mathematically explicit formulation of the latent interface, including deterministic quaternion unit projection
(Eq.~\eqref{eq:unitproj}) and mutual-information-based decoupling objectives (Eq.~\eqref{eq:mi_hinge}) intended to limit cross-talk between orientation and non-orientation
factors. Together, these design choices support a coherent explanation for the observed improvements in convergence speed: the policy learns in a lower-dimensional,
task-relevant manifold and can (when enabled) condition directly on environment parameters rather than implicitly treating them as unmodeled stochasticity.
In the runs reported here, the best-performing and most stable learning curve is obtained with the Latent+Env variant (Run~A).Future work will focus on extending the present open-loop setting to hybrid or fully closed-loop execution (intermittent vision and/or tactile feedback), expanding evaluation to broader object/gripper morphology variation and systematic OOD sweeps over environment regimes, strengthening baseline coverage with full multi-seed reporting, and
validating on physical hardware to characterize sim-to-real performance under realistic sensing and contact uncertainties. However, because we report representative single-seed learning curves, further validation would be needed for a universal improvement from explicit $e$ with multi-seed statistics and targeted OOD context sweeps.

\section*{Acknowledgments}
At this point we would like to acknowledge the usage of generative AI for the grammatical enhancement of parts of this draft manuscript. In addition, we would like to note that we are at the moment scaling up the infrastructure for the experiments, making it capable for the simulation of grasping on over 24000 different target objects, over 1000 different grippers, and with fully controllable high-fidelity physical representation of the different environmental parameters, all of which aim to further validate our assumptions about error bounds and the faster adaptation rates reported in our previous work~\cite{askianakis2024grammarizationbasedgraspingdeepmultiautoencoder}.


\newpage


\bibliographystyle{unsrtnat}
\bibliography{Bibliography}


\newpage

\appendix
\section{Additional Training Dashboards}
\label{app:dashboards}

\begin{figure}[H]
  \centering
  \includegraphics[width=0.98\linewidth]{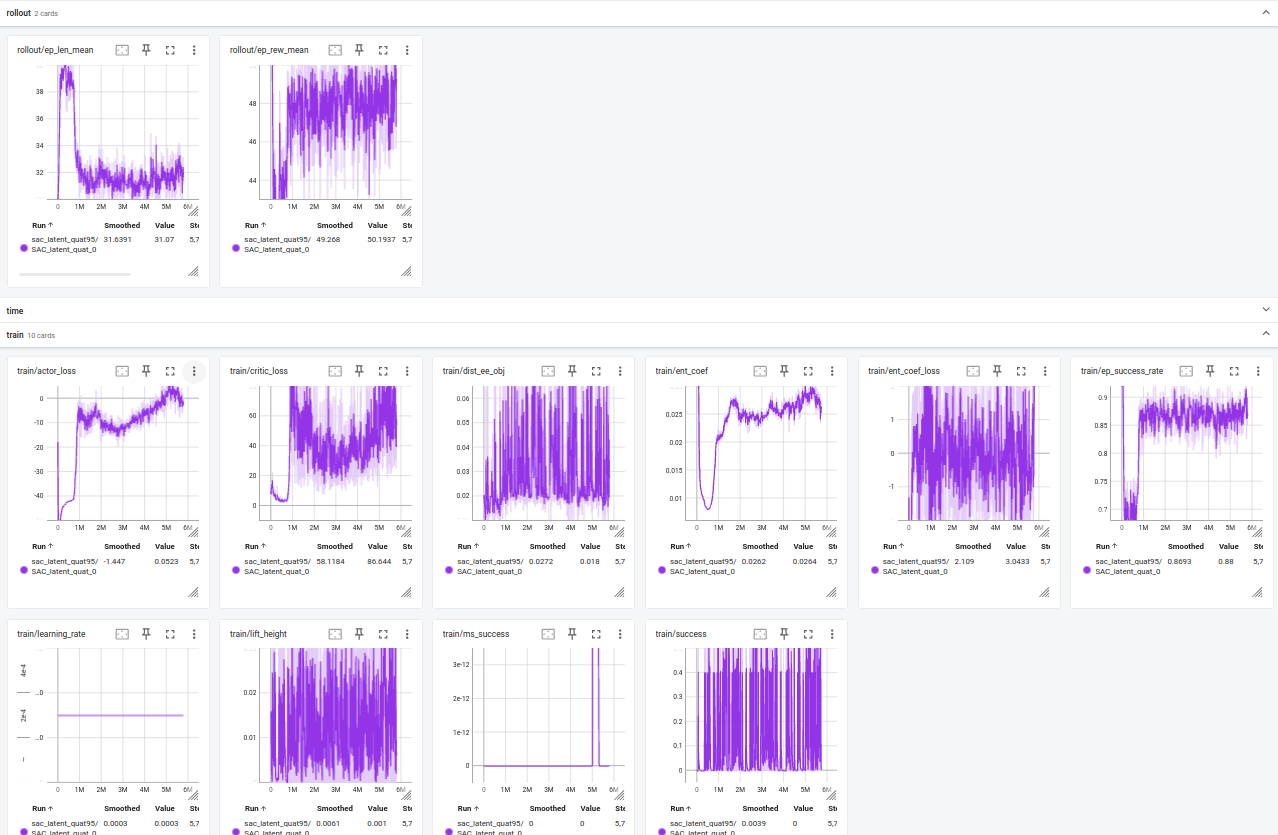}
  \caption{(Supplementary) TensorBoard dashboard for the grammarized latent policy training run, showing representative optimization metrics (reward, episode length, entropy coefficient, and auxiliary signals).}
  \label{fig:tb_latent_dashboard}
\end{figure}

\begin{figure}[H]
  \centering
  \includegraphics[width=0.98\linewidth]{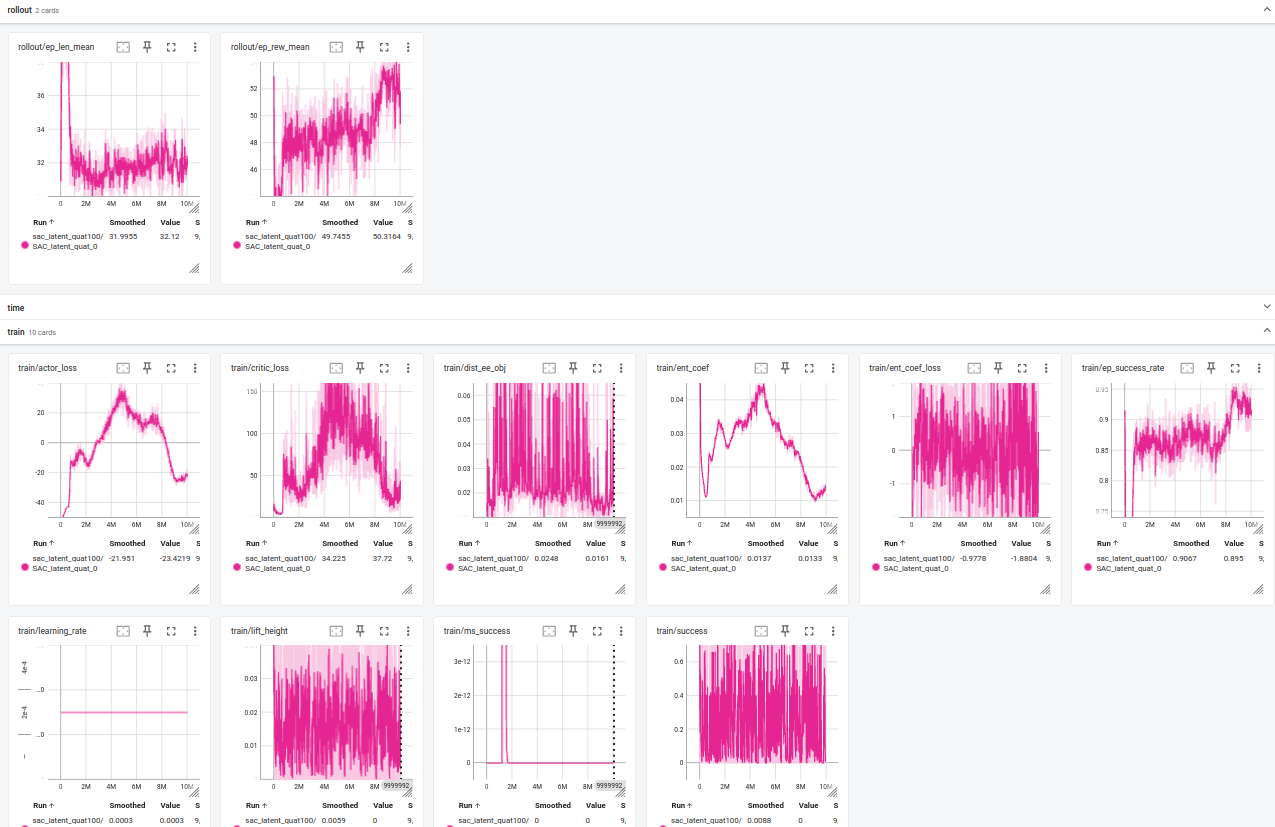}
  \caption{(Supplementary) TensorBoard dashboard for the grammarized latent+environment policy training run, showing representative optimization metrics (reward, episode length, entropy coefficient, and auxiliary signals).}
  \label{fig:tb_latent_env_dashboard}
\end{figure}

\newpage

\section{Nomenclature}

\begin{longtable*}{@{}l @{\quad=\quad} l@{}}

\(t\) & discrete time-step index within an episode \\[0.15em]
\(H\) & episode horizon (number of control steps) \\[0.15em]
\(x_0\) & single-shot exteroceptive observation at episode start \\[0.15em]
\(x_T,\;x_G\) & target/scene input and gripper descriptor (one-shot) \\[0.15em]
\(p_t\) & low-dimensional geometric term (e.g., relative end-effector--to--object position) \\[0.15em]

\addlinespace

\(z_T,\;z_G\) & modality embeddings: \(z_T = E_T(x_T)\), \(z_G = E_G(x_G)\) \\[0.15em]
\(z\) & grammarized latent representation, \(z=[z_q\Vert z_s]\) \\[0.15em]
\(z_C\) & control latent code, \(z_C=[z_q\Vert z_s\Vert e]\) \\[0.15em]
\(z_q\) & quaternion/orientation latent block, \(z_q\in\mathbb{R}^{4}\) \\[0.15em]
\(z_s\) & non-orientation latent block (shape/scene/gripper factors) \\[0.15em]
\(d_z,d_s,d_e\) & dimensions of \(z\), \(z_s\), and environment context \(e\) \\[0.15em]

\addlinespace

\(q\) & unit quaternion (orientation), \(q\in\mathbb{S}^3\subset\mathbb{R}^4\) \\[0.15em]
\(\hat q\) & raw (pre-projection) quaternion in \(\mathbb{R}^4\) \\[0.15em]
\(q_{\mathrm{ref}}\) & reference quaternion for sign-consistent projection \\[0.15em]
\(\mathcal{P}(\cdot)\) & unit-norm quaternion projection operator (Eq.~\eqref{eq:unitproj}) \\[0.15em]
\(\epsilon\) & small constant for numerical stability in projection \\[0.15em]
\(d_{\mathrm{geo}}(\cdot,\cdot)\) & geodesic orientation error on \(\mathrm{SO}(3)\) (Eq.~\eqref{eq:geodesic}) \\[0.15em]

\addlinespace

\(\tilde e\) & raw/unnormalized environment parameter vector \\[0.15em]
\(e\) & normalized environment/context vector, \(e\in[-1,1]^{d_e}\) \\[0.15em]
\(\phi_i(\cdot)\) & per-coordinate transform used in environment normalization (Eq.~\eqref{eq:env_norm_method}) \\[0.15em]
\(\mathrm{clip}(\cdot)\) & elementwise clipping to a specified interval \\[0.15em]
\(\boldsymbol{\xi}\) & episode-wise randomized physics parameter vector (Eq.~\eqref{eq:theta_def}) \\[0.15em]
\(\mu_{\mathrm{obj}},\mu_{\mathrm{table}},\mu_{\mathrm{gripper}}\) & friction coefficients (object/table/gripper) \\[0.15em]
\(m_{\mathrm{scale}}\) & object mass scaling factor \\[0.15em]
\(g_z\) & gravity acceleration along vertical axis \\[0.15em]
\(c_{\mathrm{rest}}\) & coefficient of restitution \\[0.15em]
\(d_{\mathrm{lin}}, d_{\mathrm{ang}}\) & linear and angular damping coefficients \\[0.15em]

\addlinespace

\(a_t\) & continuous control action at time \(t\) \\[0.15em]
\(a=(\Delta z_q,\Delta z_s)\) & latent action update (when acting in latent-space blocks) \\[0.15em]
\(\pi_\theta\) & parameterized policy (SAC actor) \\[0.15em]
\(r(\cdot)\) & per-step reward \\[0.15em]
\(\gamma\) & RL discount factor \\[0.15em]
\(\alpha\) & SAC entropy temperature (learned entropy coefficient) \\[0.15em]
\(\eta_{\mathrm{lr}}\) & optimizer learning rate (e.g., SAC update step size) \\[0.15em]

\addlinespace

\(\mathrm{succ}^{(k)}\) & episode-level success indicator for episode \(k\) \\[0.15em]
\(S(n)\) & rolling success rate reported at environment step \(n\) \\[0.15em]
\(\tau\) & target success threshold in time-to-threshold metric (Eq.~\eqref{eq:time_to_threshold}) \\[0.15em]
\(W\) & sustain window length for threshold crossing (Eq.~\eqref{eq:time_to_threshold}) \\[0.15em]
\(N_{\tau}\) & time-to-threshold environment steps (Eq.~\eqref{eq:time_to_threshold}) \\[0.15em]

\addlinespace

\(E_T,E_G\) & target and gripper encoders \\[0.15em]
\(E_3,D_3\) & fusion encoder/decoder (AE\(_3\)) \\[0.15em]
\(A,c\) & quaternion affine map parameters, \(z_q=Aq+c\) \\[0.15em]
\(W_s,b_s\) & fusion weights/bias in \(z_s\) encoder (Eq.~\eqref{eq:fusion_enc}) \\[0.15em]
\(\hat e\) & reconstructed environment vector from the fusion decoder \\[0.15em]
\(\mathcal{L}_{\mathrm{AE3}}\) & fusion autoencoder reconstruction objective (Eq.~\eqref{eq:ae3_loss}) \\[0.15em]
\(\lambda_q,\lambda_E,\lambda_r\) & loss weights in \(\mathcal{L}_{\mathrm{AE3}}\) \\[0.15em]
\(\varepsilon_{\mathrm{rec}}\) & representation reconstruction error (held-out; used in theory) \\[0.15em]

\addlinespace

\(\widehat I_{\mathrm{NCE}}\) & InfoNCE estimate of mutual information (Eq.~\eqref{eq:infonce_method}) \\[0.15em]
\(I_{\max}\) & mutual-information ceiling in hinge regularizer (Eq.~\eqref{eq:mi_hinge}) \\[0.15em]
\(\beta\) & hinge weight for MI regularizer (Eq.~\eqref{eq:mi_hinge}) \\[0.15em]
\(\tau_{\mathrm{NCE}}\) & InfoNCE temperature parameter (Eq.~\eqref{eq:infonce_method}) \\[0.15em]

\addlinespace

\(L_r, L_P\) & reward and transition Lipschitz constants (Assumption A2) \\[0.15em]
\(R_{\max}\) & bound on per-step reward magnitude \\[0.15em]
\(\mathrm{TV}(\cdot,\cdot)\) & total variation distance \\[0.15em]
\(\mathrm{KL}(\cdot\|\cdot)\) & Kullback--Leibler divergence \\[0.15em]
\(W_1(\cdot,\cdot)\) & Wasserstein-1 distance \\[0.15em]

\addlinespace

\(_0\) & quantity evaluated at episode start (single-shot) \\[0.15em]
\(_t\) & time-step index within an episode \\[0.15em]
\(_{\min},\,_{\max}\) & lower/upper bounds used for normalization or sampling \\[0.15em]
\(_{\mathrm{rec}}\) & quantity related to reconstruction error \\[0.15em]

\end{longtable*}

\end{document}